\documentclass{article} % For LaTeX2e
\usepackage{colm2024_conference}
\colmfinalcopy

\usepackage{microtype}
\usepackage{hyperref}
\usepackage{url}

\usepackage{bm}
\definecolor{darkred}{rgb}{0.65, 0.11, 0}

\usepackage{multicol}
\usepackage{multirow}
\usepackage{booktabs}

\usepackage{xspace}
\usepackage{graphicx}
\usepackage{pgfplots}
\usepackage{subcaption}
\usepackage{pgfplotstable}
\usepgfplotslibrary{external}
\usepackage{tikz}
\usetikzlibrary{positioning, shapes.geometric}
\usetikzlibrary{plotmarks}
\usetikzlibrary{arrows.meta}
\usepgfplotslibrary{polar}
\usepackage{amsmath}
\usepackage{dsfont}
\usepackage{amsfonts}
\pgfplotsset{compat=1.17}
\newcommand{\ourmethod}{\textsc{FlexTaF}\xspace}
\newcommand{\ourmethods}{\textsc{FlexTaF}-Single\xspace}
\newcommand{\ourmethodv}{\textsc{FlexTaF}-Vote\xspace}
\definecolor{cpurple}{rgb}{0.675, 0.573, 0.922}
\definecolor{cblue}{rgb}{0.310, 0.757, 0.910}
\definecolor{cgreen}{rgb}{0.310, 0.563, 0.214}
\definecolor{corange}{rgb}{1, 0.808, 0.329}
\definecolor{cred}{rgb}{0.8, 0.3, 0.3}
\definecolor{single_light}{rgb}{1, 0.949, 0.8}
\definecolor{single}{rgb}{0.902, 0.569, 0.220}
\definecolor{vote_light}{rgb}{0.788, 0.855, 0.973}
\definecolor{vote}{rgb}{0.043, 0.325, 0.580}

\title{\ourmethod: Enhancing Table Reasoning with Flexible Tabular Formats}

\author{Xuanliang Zhang$^{1}$, Dingzirui Wang$^{1}$, Longxu Dou$^{1}$, Baoxin Wang$^{2}$, Dayong Wu$^{2}$\\
\textbf{Qingfu Zhu$^{1}$, Wanxiang Che$^{1}$}\\
$^{1}$Harbin Institute of Technology, $^{2}$iFLYTEK Research\\
\texttt{\{xuanliangzhang, dzrwang, lxdou, qfzhu, car\}@ir.hit.edu.cn} \\
\texttt{\{bxwang2, dywu2\}@iflytek.com}
}

\begin{document}
    \maketitle
    
    \begin{abstract}
    % 表格推理任务旨在根据提供的表格对用户的query进行推理并回答
    The table reasoning task aims to answer the question according to the given table. 
    % 使用LLM已经是表格推理任务的主流方法
    Currently, using Large Language Models (LLMs) is the predominant method for table reasoning.  
    % 输入LLM的表格表示会对LLM的表格推理性能产生影响，然而前人工作通常固定一种表格表示，可能会限制模型性能
    Most existing methods employ a fixed tabular format to represent the table, which could limit the performance.
    % However, we argue that using a fixed tabular format could limit the performance, where we claim that different instances and models have flexible most suitable tabular formats. 
    % 因为解决不同的instance需要不同方面的能力，而模型不同方面的能力也各有不同，所以我们论证不同的instance和模型适合的表格表示不同
    % Because solving each instance requires different capabilities, and the capabilities of different models are also different, we claim that different instances and models suit different tabular formats. 
    Given that each instance requires different capabilities and models possess varying abilities, we assert that different instances and models suit different tabular formats.
    % 对于第一个方面，我们从实验结果的角度进行定量分析证明了我们的论断
    We prove the aforementioned claim through quantitative analysis of experimental results, where different instances and models achieve different performances using various tabular formats. 
    % 对于第二个方面，我们提出了两个方法，Single方法通过先在每个instance上预测模型最适用的表格y表示，再用预测的表示将表格输入；Vote方法则需要集成不同表示的结果
    % 基于上述讨论，我们提出了s和v，通过使用不同的表格表示来增强表格推理的性能
    Building on this discussion, we propose \ourmethods and \ourmethodv to enhance table reasoning performance by employing flexible tabular formats. 
    Specifically, (i) \ourmethods trains a classifier to predict the most suitable tabular format based on the instance and the LLM. 
    (ii) \ourmethodv integrates the results across different formats. 
    % 我们在WikiTableQuestion和TabFact两个数据集上进行实验，相比较单一表示的最好性能，我们的方法平均提升了%
    % We conduct experiments on WikiTableQuestion and TabFact and get the results with an average of \% compared with the best performance with a single tabular format, proving the effectiveness of our methods.
    % We conduct experiments on WikiTableQuestions and TabFact, which shows significant improvement with an average of 2.3\% and 4.8\% compared with the best performance of a fixed tabular format with greedy decoding and self-consistency decoding, proving the effectiveness of our methods\footnote{Our code and data will be released upon the acceptance.}. 
    Our experiments on WikiTableQuestions and TabFact reveal significant improvements, with average gains of $2.3\%$ and $4.8\%$ compared to the best performance achieved using a fixed tabular format with greedy decoding and self-consistency decoding, thereby validating the effectiveness of our methods\footnote{Our
code and prompts are available at \url{https://github.com/zhxlia/FLEXTAF}.}. 
    \end{abstract}

    \section{Introduction}
        % 表格推理，旨在帮助人们从表格中自动抽取并推理出想要获得的信息，是一项重要的自然语言处理任务
Table reasoning is a crucial task in natural language processing that aims to automatically extract and infer information from tables \citep{dong2022tablesurvey-plm}.
% 具体地，表格推理任务需要模型根据表格完成用户需求。
% Specifically, the table reasoning task requires the model to answer the question according to the table, which we collectively refer to as an instance. 
In this task, a model needs to answer questions based on the information contained in the table, with each question-table pair referred to as an instance.
% 我们把用户问题和表格统称为instance
% We collectively refer to the user question and the table as an instance. 
% 随着LLM的发展，LLM展现出了优越的常识推理和逻辑推理的能力，所以研究者尝试将LLM用于表格推理任务
Given the superior commonsense and logical reasoning capabilities of Large Language Models (LLMs),  
% especially with the Chain-of-thoughts (CoT)~\citep{chain-ot-thought} and Program-of-Thoughts (PoT) prompts~\citep{pot,pal}, 
researchers increasingly utilize them for table reasoning, which has become the mainstream method \citep{chen-2023-few-one,zhang2024surveytablereasoning,dong2024tablesurvey}. 
% LLM无需微调性能就超越了微调后的模型，已成为目前的主流方法
% 因此，本文关注如何用LLM解决表格推理问题
Therefore, we focus on how to solve the table reasoning task with LLMs in this paper.

\begin{figure}[t]
    \centering
    \includegraphics[width=0.5\linewidth]{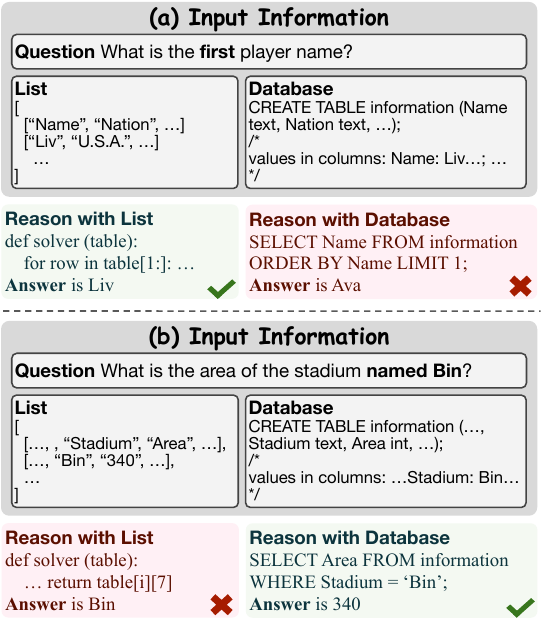}
    \caption{
    % 使用不同模型和不同表格表示的表格推理的performance
    The table reasoning performance varies with different tabular formats. 
    % List格式不利于模型查找满足特定条件的列，而Database格式不方便按顺序索引
    % The Database format is not convenient for sequential indexing, while the List format hinders the search for columns that meet specific conditions.
    The List format is convenient for sequential indexing, while the Database format facilitates the search for columns that meet specific conditions.
    }
    \label{fig:intro}
\end{figure}

% 一些之前的工作通过设计prompt来增强模型的表格推理性能，比如Chain-of-Table，通过提示模型迭代使用自然语言和代码推理来解决表格推理任务
Some previous works enhance the table reasoning performance by designing prompts \citep{cheng2023binding,dater,zhang-2024-reactable,lee2024reduce_table}, such as Chain-of-Table~\cite{Wang2024chainoftable}, which solves table reasoning tasks by prompting the LLM to iteratively reason with natural language and programs.
% 前人工作通过融合不同的推理结果来增强表格推理的性能，例如Chain-of-Table，通过结合自然语言和代码推理来解决表格推理任务
% Previous works \citep{dater,zhang-2024-reactable,liu-etal-2024-rethinking} enhance the table reasoning performance by using different reasoning formats, such as Chain-of-Table~\cite{Wang2024chainoftable} solves the table reasoning task by integrating natural language and code reasoning. 
% 然而，上述方法在输入中提供的是固定的表格形式，但前人工作表明，输入的表格格式也会影响性能
While the above methods provide a fixed tabular format in their prompts, \citet{TableMeetsLLM,singha2023tabular-representation,deng-etal-2024-tables-texts-images} argue that different table reasoning tasks have different most suitable tabular formats. 
% 然而，前人工作只关注到了表格表示会影响不同任务的性能，但解决不同的问题需要的能力不同，比如解决有的问题需要某种常识推理能力，解决有的问题需要某种数值推理能力
However, \textbf{\textit{existing works primarily analyze from the task aspect}} without considering that solving different instances demands distinct reasoning skills \cite{shi-etal-2020-squall,pot,cheng2023binding,liu-etal-2024-rethinking,chen2024sheetagent}. 
% For instance, solving one instance necessitates some common sense reasoning ability, and solving another instance requires some numerical reasoning ability. 
% 比如，解决图中例子(a)需要顺序索引的能力，而解决例子(b)需要查找满足特定条件的列的能力，database格式有助于此
% For example, as shown in Figure~\ref{fig:intro}, solving instance (a) requires reasoning with indexing sequentially which the List format is conducive to, while solving instance (b) requires the ability to identify columns that meet specific conditions that can benefit from the Database format.
For example, as shown in Figure~\ref{fig:intro}, solving instance (a) involves reasoning with sequential indexing, which is facilitated by the List format. 
In contrast, solving instance (b) requires identifying columns that satisfy specific conditions, making the Database format more suitable.
% 所以，在同一表格推理任务中对所有的instance使用相同的表格表示也会限制性能
Therefore, using a fixed tabular format for all instances could limit performance. 
Additionally, models vary in their reasoning capabilities \citep{cao-etal-2023-api,bhandari2024robustness,zhang2024benchmarkingtexttosql}, therefore the most suitable tabular format could differ for each model.

% 基于上述讨论，我们主要研究表格表示对大模型表格推理性能的影响，包括以下两点：
% Building on the above discussion, we focus on how tabular format affects the table reasoning performance of LLMs from the following two aspects.
Based on the above discussion, we focus on the impact of tabular formats on the table reasoning performance of LLMs from the following two aspects.
% 我们证明使用不同模型，面对不同表格和问题，最适用于模型求解的表格表示不同
(\emph{i})~We claim that \textbf{\textit{different instances and different LLMs require distinct most suitable tabular formats\footnote{We refer to the suitable tabular format as the one that enables the model to correctly solve a given instance.}}}.
% tabular format most suitable for LLM reasoning is relevant to the specific instance and the LLM.
% 我们提出我们的方法，分别通过预测最适合模型求解的表格表示，和集成多表格表示的结果，来提升模型性能
(\emph{ii})~We propose to enhance the table reasoning performance by \textbf{\textit{predicting the most suitable format or assembling the results from multiple tabular formats}}.

% 首先，我们讨论了“不同的instance和LLM最适合的表格表示是不同的”
First of all, we discuss that distinct tabular formats are suitable for different instances and LLMs.
% 我们使用不同的表格表示和不同的模型进行了探索性试验
We conduct exploratory experiments utilizing different tabular formats and LLMs on table reasoning datasets.
% 实验结果表明，表格推理的性能会随着表格表示和模型的变化也会有较大的变化
The experimental results present that table reasoning performance varies significantly with different formats and LLMs.
% 基于以上观点，我们提出了我们的方法，Single和Vote，基于不同表示的结果来增强表格推理的性能”
Based on the above analysis, we propose \ourmethod, which includes \ourmethods and \ourmethodv, to enhance table reasoning performance through flexible tabular formats.
% Single方法通过训练分类器来基于instance和LLM选取最合适的表格表示
\ourmethods identifies the most suitable tabular format based on the instance and the LLM by training the classifier.
% Vote方法将多个表格表示的结果进行投票来选出最终的答案
\ourmethodv determines the final answer by voting on results obtained from multiple formats. 
% 比较之下，
In comparison, although \ourmethods requires training data but only infers once, \ourmethodv is training-free but with expensive inference costs.
% training-free but with expensive inference cost;一个是 require training data but only inference once.

% 为了证明我们方法的有效性，我们分别在WikiTableQuestions(简称为WikiTQ)，TabFact这两个主流表格推理数据集上进行实验
To demonstrate the effectiveness of our methods, we conduct experiments on two mainstream table reasoning datasets: WikiTableQuestions~\citep{pasupat-liang-2015-WikiTableQuestions} and TabFact~\citep{2019TabFact}.
% 我们的方法相比较同一模型在该数据集上性能最好的表格表示的结果，分别平均提升%，证明了我们方法的有效性
Compared to the best performances achieved by using a fixed format with greedy decoding and self-consistency decoding~\cite{wang2023selfconsistency}, \ourmethods and \ourmethodv show average improvements of $2.3\%$ and $4.8\%$ respectively with comparable inference costs, confirming the effectiveness of our methods.
% 同时，进一步的分析实验表明，不同的表格表示都有自己能做对的instances，证明了我们“不同instance最合适的表格表示是不同的”的观点
Further analysis experiments reveal that some instances can only be correctly solved using a specific format, proving that the most suitable tabular formats for different instances are distinct.

% 我们的贡献如下
Our contributions are as follows:
\begin{enumerate}
    % 我们理论上证明了使用不同模型进行表格推理，面对不同问题和表格，最适合于模型求解的表格表示不同
    \item We claim that the most suitable tabular formats for different instances and LLMs are distinct.
    % 我们提出了两种方法，通过在instance的粒度上变化表格表示来提高模型表格推理的性能
    \item We propose \ourmethod, which includes \ourmethods and \ourmethodv, to enhance table reasoning performance by utilizing flexible tabular formats.
    % 实验结果表明，我们的方法相比较单一表示的最好结果，平均提升%，证明了我们方法的有效性
    \item Our experimental results indicate that \ourmethods and \ourmethodv achieve average improvements of $2.3\%$ and $4.8\%$ respectively, over the best results obtained using fixed formats with greedy decoding and self-consistency decoding at comparable inference costs, demonstrating the effectiveness of our methods.
\end{enumerate}

    \section{Discussion on Impact of Tabular Formats}
        \label{sec:analysis}
        % 在本节，我们将从实验结果的角度证明不同模型，不同问题和表格，最适合求解的表格表示不同
We first claim that the most suitable tabular formats for different instances and LLMs are distinct. 
% 因为求解不同的instance需要不同方面的能力，因此需要对应不同的表格表示来求解
Since resolving different instances requires diverse abilities \cite{pot,cheng2023binding,chen2024sheetagent}, it is necessary to tailor tabular formats for each instance accordingly. 
% 并且不同模型的能力也有差异，导致不同模型适用的表格表示也不同
Additionally, the capabilities of different models vary \citep{cao-etal-2023-api,bhandari2024robustness,zhang2024benchmarkingtexttosql}, resulting in different tabular formats suitable for different LLMs. 
% 为了进一步证明我们的观点，在本节，我们从实验结果的角度分析
We further discuss from the perspective of experimental results in this section.

We select five popular formats, including Markdown, Dict, List, Pandas, and Database formats, following previous works \cite{singha2023tabular-representation,cheng2023binding,Wang2024chainoftable,liu-etal-2024-rethinking}. 
% 我们在附录中详细介绍这5种表示
Detailed descriptions of these formats are provided in the supplementary material.
% 下面，我们将从模型、问题和表格两个方面定量地分析差异
Subsequently, we quantitatively analyze how tabular format affects the table reasoning performance of LLMs from the aspects of the instance and the model respectively.

\subsection{The Relationship between Instance and Tabular Format}
    % 下面我们讨论，不同的instance最适合的表格表示也是不同的
    We first claim that \textit{\textbf{different instances suit different tabular formats}}.
    % 表1展现了针对不同的表格表示能做对的instance中，只有该表格表示能做对的instance的比例
    Table~\ref{tab:analysis_example} presents the percentage of instances that can only be correctly resolved by each tabular format.
    % 从表中可以发现，在固定模型的情况下，不同的表格表示都有只有该表示能做对的instance，甚至对于某些表格表示，其能专门做对的instance的比例超过了20%
    It can be observed that, with the same fixed LLM, each tabular format is uniquely suited to certain instances, with some formats correctly solving over $20\%$ of the instances exclusively.
    % 上述结果表明，在固定模型的情况下，针对不同的instance，其最合适的表格表示形式也是不同的
    The results indicate that, for a given LLM, the most suitable tabular formats vary according to the specific instances.

\begin{figure*}[t]
    \centering
    \includegraphics[width=0.8\linewidth]{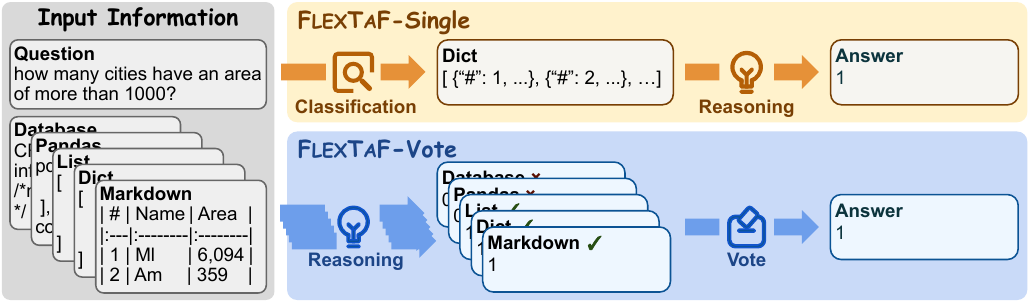}
    \caption{
    % 我们方法的流程图
    The overview of \ourmethod. 
    \ourmethods consists of two steps: 
    (\emph{i})~Classification: A classifier we trained predicts the most suitable tabular format based on the given instance and model. 
    (\emph{ii})~Reasoning: Using the predicted format, the LLM solves the instance by representing the table accordingly.
    \ourmethodv consists of two steps: 
    % 我们的Vote方法用多种表格表示分别推理，再在得到的结果上实施投票
    (\emph{i})~Reasoning: Various formats are employed to represent the table and facilitate reasoning with the LLM, resulting in multiple answers.
    (\emph{ii})~Vote: The final answer is determined using a voting mechanism.
    }
    \label{fig:method}
\end{figure*}

% 模型和表格表示的关系
\subsection{The Relationship between Model and Tabular Format}
    % 我们首先证明表格表示对性能的影响与使用的模型有关。
    In this subsection, we discuss that \textit{\textbf{different models suit different tabular formats}}. 
    % 表1展示了不同的模型对应的性能最好的的表格表示是不同的，并且不同表格表示带来的性能波动很大
    Table~\ref{tab:main} demonstrates that the most suitable tabular format varies across models and the performance gap arises due to different formats. 
    % 比如，Llama3在使用markdown这种表格表示时的性能显著优于其他表格表示，而DeepSeek-Coder在使用JSON、DB时的性能又会优于使用md
    For instance, Llama3~\cite{Llama3} exhibits significantly better performance with the Markdown format compared to other formats, while DeepSeek-Coder~\cite{Deepseek-Coder} performs better with Dict and Database formats than with Markdown.
    % 我们采用卡方检验证明了不同模型上的正确表格表示的分布存在显著差异
    We employ a Chi-square test to demonstrate significant differences in the distribution of the most suitable tabular formats across different models, as detailed in the supplementary material.

    \section{Methodology}
        \label{sec:methodology}
        % 在本节，我们详细介绍我们的两种方法。
In this section, we introduce \ourmethod, which consists of \ourmethods and \ourmethodv.
% 本方法的overview见图2
The overview of \ourmethod is shown in Figure~\ref{fig:method}.

\subsection{Task Definition}
% 本方法主要关注于表格推理任务，可以形式化定义如下
\ourmethod focuses on the table reasoning task, which can be formally defined as follows: 
% 给定用户requirement Q，和表格T，输入模型M
Given an instance $I$ comprising a natural language question $Q$ and a table $T$, the table reasoning task aims to derive the corresponding answer $\hat{A} = \texttt{M}(\texttt{F}(T), Q)$,
% 其中Ri为表格表示，D为demonstrations
 where $\texttt{M}$ represents the model and $\texttt{F}$ denotes the tabular format. 
% 为了更好地解决表格推理任务，A为正确答案A*的概率应该尽可能大，即 
To effectively solve the table reasoning task, the probability that the generated answer $\hat{A}$ matches the gold answer $A^*$ should be maximized.

\subsection{\ourmethods}
% 基于第二章的讨论，需要找到最合适的表格表示，因此\ourmethods通过训练classifier来找到最合适的表格表示
As discussed in \S\ref{sec:analysis}, it is essential to determine the most suitable tabular format, so \ourmethods achieves this by training a classifier to identify the most suitable format.

\subsubsection{Classification}
Classification aims to predict the most suitable tabular format from a set of candidate formats based on the instance and the model.
% 形式化表述为
It can be formally expressed as $\hat{\texttt{F}} = \texttt{CLS}_\texttt{M}(I), \hat{\texttt{F}} \in \texttt{F}_{\texttt{M}, I}^* = \{\texttt{F} | \texttt{M}(\texttt{F}(T), Q) = A^*\}$.

\paragraph{Training Data Collection}
% 为了令分类器能够根据问题和表格，预测最适合模型推理的表格表示，我们用模型M对训练集中的数据进行标注。
To train the classifier to predict the most suitable tabular format for the given instance $I$ and the LLM $\texttt{M}$, we annotate the training data with $\texttt{M}$. 
% 具体来说，对于训练集中的每个instance，我们分别用不同的表格表示进行推理，并评估不同表示得到的结果的正确性。
Specifically, for each instance in the training set, we utilize all candidate tabular formats to reason respectively and evaluate the correctness of each answer. 
% Specifically, for each instance in the training set, we assess all candidate tabular formats by reasoning through each and evaluating the correctness of the resulting answers. 
% 因此，我们收集到了每个instance中，模型M对于问题和表格能做对的表格表示，即，作为分类器的训练集。
We then collect the set of most suitable tabular formats for each instance in the training data, denoted as $\{\texttt{F}_{\texttt{M}, I_t}^*\}$, for which $\texttt{M}$ can correctly reason with the format on the instance $I_t$, and use these as the training data for our classifier. 
% 并且，我们采取了数据过滤策略，指我们移除了训练数据中有一半以上表格表示都做对的instance，因为这些instance不能很好地体现不同表格表示之间的区别
Additionally, we take a data filtering strategy to remove instances from the training set where more than half of the candidate formats are correct or where none are correct, as such instances do not effectively highlight differences between the tabular formats.

\paragraph{Learning Objective}
% 由于$R_{M, I}^*$的数量可能不止一个，所以我们采用多标签分类的训练方法，其中每个标签代表对于这个instance推理正确的表格表示。
Since there could be multiple formats in $\texttt{F}_{\texttt{M}, I}^*$, we apply a multi-label classification training method \citep{Binary-relevance-MLC,tsoumakas2007multi_lable_classification_overview,multi-label-classify-book}, where each label denotes a tabular format. 
% 我们采用binary relevance多标签分类方法，将所有表格表示序列化，为每个instance的对应的表格表示二值化。
% We employ the binary relevance multi-label classification method~\citep{Binary-relevance-MLC}, which involves serializing all tabular formats and binary formats for each instance. 
Moreover, we utilize a binary relevance method~\citep{Binary-relevance-MLC} for multi-label classification. 
Specifically, the tabular formats are serialized and transformed into binary vectors for each instance.
% 训练中，我们采用二元交叉熵损失函数，即
During training, we adopt Binary Cross-Entropy Loss, normalized by the number of instances $N$, as follows:
\begin{equation}
    \texttt{L}(\hat{y}, y) = -\frac{1}{N}\sum_{i=1}^{N}\sum_{r=1}^{|\texttt{F}|}[y_{ir}\texttt{log}(\hat{y}_{ir}) + (1-y_{ir})\texttt{log}(1-\hat{y}_{ir})]
\end{equation}
% 其中，R指所有表格表示的数量，yir指样本i在表格表示r上的真实标签，即0或1，\hat{y}_{ir}代表样本i在r上的预测概率
Among them, $|\texttt{F}|$ refers to the total number of candidate tabular formats, $y_{ir}$ indicates the gold value for the instance $i$ on tabular format $r$, with possible values of $0$ or $1$, and $\hat{y}_{ir}$ represents the predicted probability for instance $i$ on the tabular format $r$.

\paragraph{Predicting Tabular Format}
% 得到分类器后，我们根据分类器预测的表格表示进行表格推理
After obtaining the classifier, we predict the most suitable tabular format $\hat{\texttt{F}}$.
% 我们正则化所有标签的预测分数，并选取其中概率最大的表示作为我们的分类结果。
We regularize the predicted scores of all formats and select the format with the highest probability as our classification result $\hat{\texttt{F}}$.

\subsubsection{Reasoning}
% 在预测最合适的表格表示后，我们用预测的表格表示进行推理
After predicting the most suitable table format $\hat{\texttt{F}}$, we utilize the predicted format for table reasoning.
% 具体地，我们用预测的format表示表格，和问题、示例一同输给LLM，得到结果
Specifically, we represent the table with the predicted format $\hat{\texttt{F}}$ in the prompt and employ the LLM to derive the final answer.

\subsection{\ourmethodv}
% \ourmethods需要人工标注数据，可能难以获取，因此我们提出\ourmethodv，通过集成不同表示的结果来得到最终的答案
\ourmethods requires manual data annotation, which could be difficult to obtain, so we propose \ourmethodv to obtain the final answer by integrating the results from multiple tabular formats, inspired by \citet{wang2023selfconsistency,qin-etal-2023-cross-lingual-prompting,luo2024multipot}.

\subsubsection{Reasoning}
% 具体地说，我们分别用多种表示表达表格，将表格、问题输入LLM，导致模型的推理路径不同，可能得到多个不同的答案。
We first construct multiple table reasoning prompts, representing the table with different formats in each prompt. 
Then we employ these prompts to reason with the LLM, obtaining multiple answers accordingly.

\subsubsection{Vote}
% 然后，我们采用投票机制保留在多个结果中最一致的结果，可以形式化表示为
We retain the most consistent result across multiple results by adopting a voting mechanism, which can be formally expressed as the following equation.
\begin{equation}
    \hat{A} = \texttt{argmax}_A\sum_{r=1}^{|\texttt{F}|} \mathds{1}(A_r=A)
\end{equation}
% 这里，｜A｜是所有可能的答案数，Ai是不同表格表示带来的不同答案，A代表所有可能的答案
% 函数返回1如果f为True，否则返回0。
Here, $|\texttt{F}|$ is the total number of candidate formats, $A_r$ is the answer obtained from the tabular format $\texttt{F}_r$, and $A$ is each possible answer. 
The function $\mathds{1}(f)$ returns $1$ if $f$ is true and $0$ otherwise. 
% 如果有多个相同数量的答案，如果采用开源模型进行实验，我们选取其返回的log probabilities最大的答案，如果采用闭源模型，我们从中随机选取一个。
In the event of a tie, we select the answer with the highest logarithmic probability following \citet{luo2024multipot}.

\subsection{Comparison}
% 为了更好地运用我们的两种方法，我们比较这两种方法的适用场景
To better employ our two methods, we examine their application scenarios.
% 1. \methods适用于对在线推理的效率要求高的场景，但需要标注数据并事先训练模型
(\emph{i})~\ourmethods is ideal for scenarios where high-efficiency online reasoning is required, although it necessitates prior training data annotation and classifier training.
% 2. \methodv适用于缺少标注数据的场景，对实时推理效率有一定容忍性
(\emph{ii})~\ourmethodv suits scenarios where annotated data is unavailable while maintaining a certain tolerance for real-time reasoning efficiency.

\begin{table*}[t]
    \centering
    \small
    \begin{tabular}{l|cc|cc|cc|cc}
    \toprule
    \multirow{3}{*}{\textbf{Tabular Format}} & \multicolumn{4}{c|}{\textbf{WikiTQ}} & \multicolumn{4}{c}{\textbf{TabFact}} \\
    & \multicolumn{2}{c|}{\textbf{Llama3}} & \multicolumn{2}{c|}{\textbf{DeepSeek-Coder}} & \multicolumn{2}{c|}{\textbf{Llama3}} & \multicolumn{2}{c}{\textbf{DeepSeek-Coder}} \\
    & 8B & 70B & 6.7B & 33B & 8B & 70B & 6.7B & 33B \\
    \midrule
    Markdown & $47.7$ & $63.3$ & $32.2$ & $31.2$ & $75.2$ & $86.4$ & $60.4$ & $63.6$ \\
    Dict & $43.0$ & $56.4$ & $25.6$ & $53.6$ & $65.5$ & $80.0$ & $63.9$ & $78.0$ \\
    List & $30.5$ & $56.3$ & $19.2$ & $50.8$ & $57.4$ & $77.5$ & $63.7$ & $75.4$ \\
    Pandas & $39.2$ & $52.3$ & $39.7$ & $48.8$ & $47.8$ & $73.6$ & $62.5$ & $75.9$ \\
    Database & $31.0$ & $48.2$ & $41.8$ & $45.5$ & $65.0$ & $75.0$ & $70.7$ & $76.3$ \\
    \midrule
    % $\text{Acc}_{\text{max}} - \text{Acc}_{\text{min}}$ & $17.2$ & $15.1$ & $22.6$ & $22.4$ & $29.6$ & $12.8$ & $11.5$ & $18.0$ \\
    \ourmethods & \bm{$50.5$} & \bm{$69.1$} & \bm{$46.3$} & \bm{$54.5$} & \bm{$77.0$} & \bm{$87.1$} & \bm{$70.9$} & \bm{$78.3$} \\
    $\Delta$ & $+2.8$ & $+5.8$ & $+4.5$ & $+0.9$ & $+1.5$ & $+0.7$ & $+2.2$ & $+0.3$\\
    % \midrule
    % Oracle & $71.8$ & $83.5$ & $67.7$ & $74.6$ & $96.9$ & $98.1$ & $95.4$ & $97.1$ \\
    \bottomrule
\end{tabular}

    \caption{
    % 我们方法和使用一种表格表示的对比
    The accuracy of reasoning using a fixed tabular format with \textbf{greedy decoding} and \ourmethods, across four LLMs on WikiTableQuestions (WikiTQ) and TabFact. 
    The best performance for each LLM and dataset is marked in \textbf{bold}. 
    % delta指的是我们方法相对使用一种表示的最佳性能的提升
    $\Delta$ denotes the improvement of \ourmethods relative to the best performance of using a fixed format with greedy decoding for each LLM and dataset.
    }
    \label{tab:analysis_model}
\end{table*}

    \section{Experiments}
        \label{sec:experiments}
        \subsection{Settings}
    \subsubsection{Datasets}
    % 我们使用WikiTQ和TabFact数据集来评估我们的方法
    We use WikiTableQuestions~\citep{pasupat-liang-2015-WikiTableQuestions} and TabFact~\citep{2019TabFact} datasets to evaluate \ourmethod.
    % WikiTQ是表格问答任务的主流数据集，包含多种领域和问题类型，需要模型根据提供的表格，给出用户问题的答案。
    WikiTableQuestions is a mainstream dataset for the table Question Answering task, containing diverse questions across various domains, which requires answering the question based on the table.
    % TabFact是表格事实验证任务的主流数据集，需要模型根据表格判断断言是entailed or refuted，其中的statements包含Linguistic Reasoning和Symbolic Reasoning两种推理形式。
    TabFact is a prominent dataset for the table fact verification task, which needs to determine whether a claim is entailed or refuted by the table.

\begin{table*}[t]
    \centering
    \small
    \begin{tabular}{l|cc|cc|cc|cc}
    \toprule
    \multirow{3}{*}{\textbf{Tabular Format}} & \multicolumn{4}{c|}{\textbf{WikiTQ}} & \multicolumn{4}{c}{\textbf{TabFact}} \\
    & \multicolumn{2}{c|}{\textbf{Llama3}} & \multicolumn{2}{c|}{\textbf{DeepSeek-Coder}} & \multicolumn{2}{c|}{\textbf{Llama3}} & \multicolumn{2}{c}{\textbf{DeepSeek-Coder}} \\
    & 8B & 70B & 6.7B & 33B & 8B & 70B & 6.7B & 33B \\
    \midrule
    Markdown & $49.1$ & $62.8$ & $32.1$ & $29.7$ & $75.2$ & $86.7$ & $60.9$ & $63.6$ \\
    Dict & $44.8$ & $58.1$ & $28.7$ & $59.7$ & $67.9$ & $80.9$ & $71.9$ & $82.4$ \\
    List & $35.4$ & $59.7$ & $27.5$ & $55.6$ & $59.8$ & $78.0$ & $66.9$ & $78.0$ \\
    Pandas & $41.4$ & $56.2$ & $43.9$ & $54.4$ & $56.0$ & $74.7$ & $67.2$ & $79.2$ \\
    Database & $35.2$ & $48.6$ & $42.5$ & $46.4$ & $69.5$ & $76.0$ & $70.7$ & $77.2$ \\
    \midrule
    % \multicolumn{2}{l}{\textbf{\ourmethod}} & & \\
    % \ourmethods & $50.5$ & $69.1$ & $46.3$ & $54.5$ & $77.0$ & $87.1$ & $70.9$ & $78.3$ \\
    \ourmethodv & \bm{$55.7$} & \bm{$69.9$} & \bm{$51.4$} & \bm{$60.9$} & \bm{$80.3$} & \bm{$88.5$} & \bm{$77.9$} & \bm{$84.4$} \\
    $\Delta$ & $+ 6.6$ & $+ 7.1$ & $+ 7.5$ & $+ 1.2$ & $+ 5.1$ & $+ 1.8$ & $+ 7.2$ & $+2.0$ \\
    % Oracle & $71.8$ & $83.5$ & $67.7$ & $74.6$ & $96.7$ & $98.1$ & $95.4$ & $97.1$ \\
    \bottomrule
\end{tabular}

% \begin{tabular}{lll|ccccc|cc|c}
%     \toprule
%     \multirow{2}{*}{\textbf{Dataset}} & \multirow{2}{*}{\textbf{Model}} & \multirow{2}{*}{\textbf{Scale}} & \multicolumn{5}{c|}{\textbf{Single-Tabular-Representation}} & \multicolumn{2}{c|}{\textbf{\ourmethod}} & \multirow{2}{*}{\textbf{Oracle}} \\
%     &  &  & MD & Dict & List & PD & DB & Single & Vote & \\
%     \midrule
%     \multirow{4}{*}{WikiTQ} & \multirow{2}{*}{Llama3} & 8B & $47.7$ & $43.0$ & $30.5$ & $39.2$ & $31.0$ & $50.5$ & $55.7$ & $71.8$ \\
%     & & 70B & $63.3$ & $56.4$ & $56.3$ & $52.3$ & $48.2$ & $69.0$ & $69.9$ & $83.5$ \\
%     \cmidrule{2-11}
%     & \multirow{2}{*}{DeepSeek-Coder} & 6.7B & $32.2$ & $25.6$ & $19.2$ & $39.7$ & $41.8$ & $46.3$ & $51.4$ & $67.7$ \\
%     & & 33B & $31.2$ & $53.6$ & $50.8$ & $48.8$ & $45.5$ & $54.5$ & $60.9$ & $74.6$ \\
%     \midrule
%     \multirow{4}{*}{TabFact} & \multirow{2}{*}{Llama3} & 8B & $77.4$ & $65.5$ & $57.4$ & $47.8$ & $65.0$ & $77.0$ & $80.9$ & $96.7$ \\
%     & & 70B & $86.4$ & $80.0$ & $77.5$ & $73.6$ & $75.0$ & $87.1$ & $88.5$ & $98.1$ \\
%     \cmidrule{2-11}
%     & \multirow{2}{*}{DeepSeek-Coder} & 6.7B & $60.4$ & $63.9$ & $63.7$ & $62.5$ & $71.9$ & $70.5$ & $78.0$ & $95.4$ \\
%     & & 33B & $63.6$ & $81.6$ & $75.4$ & $75.9$ & $76.3$ & $78.3$ & $84.4$ & $97.1$ \\        
%     \bottomrule
% \end{tabular}
    \caption{
        % 单一表格表示和我们的方法在WikiTQ和TabFact上的性能
        The accuracy of reasoning using a fixed tabular format with \textbf{self-consistency decoding}~\cite{wang2023selfconsistency} and \ourmethodv, across four LLMs on WikiTQ and TabFact. 
        The best performance for each LLM and dataset is marked in \textbf{bold}. 
        $\Delta$ denotes the improvement of \ourmethodv relative to the best performance of using a fixed format with self-consistency for each LLM and dataset.
    }
    \label{tab:main}
\end{table*}

    \subsubsection{Metric}
    % 我们使用准确率作为我们的评价指标，following前人工作。
    We employ accuracy as the evaluation metric for WikiTableQuestions and TabFact, following the previous works \citep{pasupat-liang-2015-WikiTableQuestions,2019TabFact}, and use accuracy to evaluate the performance of classification.
    % 准确率衡量了模型生成正确答案的能力，只有当预测的结果和正确答案完全相同时才算生成结果正确。
    Accuracy measures the model ability to generate the correct answer, which is achieved only when the predicted result exactly matches the correct answer. 
    % 由于我们方法aims to为每个instance寻找到一个最适合模型求解的表示，所以采用accuracy衡量分类器预测的top 1是否在该表格被正确求解的表示中
    Since \ourmethods aims to identify the most suitable format, we adopt accuracy to assess whether the top format predicted by the classifier is in the suitable formats of the instance. 

    \subsubsection{Models}
    % 对于表格推理任务，我们使用Llama3，DeepSeek-Coder和gpt-3.5，对于表格表示预测任务，我们使用ELECTRA-Large
    For reasoning, we employ Llama3~\citep{Llama3} and DeepSeek-Coder~\citep{Deepseek-Coder}, and for classification, we use ELECTRA-Large~\cite{Clark2020ELECTRA}.
    % Llama3和DeepSeek-Coder因其在各种任务上优越的性能，分别是popular的general和代码开源大模型
    Llama3 and DeepSeek-Coder are popular open-source LLMs for their outstanding performance across various tasks.
    % 我们选择ELECTRA作为分类器，因为其相比较其他同等规模的预训练模型在语言理解和QA任务上展现了优越的性能。
    We choose ELECTRA-Large due to its superior performance in language comprehension and question-answering tasks compared to other pre-trained models of similar size \citep{Clark2020ELECTRA}.

    \subsubsection{Implementation Details}
    % 我们采用了前人工作中通常采用的Markdown, JSON, Data-Matrix, DFLoader, and Database5种表格表示进行实验，并且使用不同模型在不同数据集上普遍有较高性能
    For the WikiTableQuestions and TabFact datasets, we adopt Markdown, Dict, List, Pandas, and Database as tabular formats, which are commonly used in previous works \citep{singha2023tabular-representation,cheng2023binding,dater,Wang2024chainoftable} and generally have high performance across datasets with different models. 
    % 其中，对于MD，我们采取了思维链提示，而对于其他表示，我们采取Program of Thoughts提升
    To enhance table reasoning performance, we utilize the Chain-of-Thought (CoT) prompt \citep{chain-ot-thought} for Markdown and the Program-of-Thought (PoT) prompt \citep{pot,pal} for other formats.
    % 对于WikiTQ和TabFact，我们分别采用了4-shot的prompt和2-shot的prompt，因为WikiTQ中的问题更难
    In addition, we apply 4-shot prompts and 2-shot prompts respectively, since the questions in WikiTableQuestions are more challenging.
    % 为了提高WikiTQ数据集上的实验性能，我们在不同表格表示的prompt中使用的demo不完全相同，我们在4.3.1探索了在保证相同demo时不同表格表示和我们方法的性能。
    To improve the performance on WikiTableQuestions, we use different demonstrations for each tabular format, and we explore the performance of \ourmethod with unified demonstrations in \S\ref{subsub:unifying}.
    % 具体的prompt在附录1
    Detailed prompts are provided in the supplementary material. 
    % 我们为每个LLM的表格推理训练一个表格表示的分类器
    We train a tabular format classifier for each LLM on each dataset.
    % 并且为了将ELECTRA用于多标签分类任务，我们在ELECTRA输出的[CLS]标记后接5个二分类器，每个用于预测一个表格表示是否适用于求解输入的instance，具体的训练细节见附录3
    % To implement multi-label classification using ELECTRA, we append $5$ binary classifiers after the \texttt{[CLS]} token.
    We provide detailed training information in the supplementary material. 
    % 我们设置温度为0.1，因为这个温度为我们
    For sampling in self-consistency decoding~\cite{wang2023selfconsistency}, we set $temperature = 0.1$ which can bring optimal performance across most formats within $temperature \leq 0.8$.
    % 为了和我们方法比较的公平性，我们设置sampling_n=5
    To ensure a fair comparison with \ourmethodv, we set $sampling\_n=5$.

\subsection{Main Experiment}
% 我们方法和self-consistency方法的对比如表所示
Table~\ref{tab:analysis_model} and Table~\ref{tab:main} compare \ourmethods with using a fixed format with greedy decoding, and \ourmethodv with using a fixed format with self-consistency~\cite{wang2023selfconsistency}, respectively.
We observe that: 
% 1. 我们的Single方法以较小的计算资源开销，与使用单一表格表示进行self-consistency的性能可比甚至超越，平均提升%
% (\emph{i})~\ourmethods surpasses the best results achieved by the fixed format with greedy decoding by an average of $2.3\%$.
% 2. 我们的Vote方法在相同计算开销的情况下显著超越了使用单一表格表示进行self-consistency的性能，平均提升%
% (\emph{ii})~\ourmethodv significantly outperforms using the fixed tabular format with self-consistency by 4.8\% on average, with comparable computational costs. 
(\emph{i})~\ourmethods and \ourmethodv surpass the best results achieved by the fixed format with greedy decoding and self-consistency, by an average of $2.3\%$ and $4.8\%$ respectively, with comparable computational costs.
% 证明了使用统一的表格表示会限制性能，在某种表示不适合用于模型求解这个问题时，即使使用self-consistency也很难正确解决
(\emph{ii})~Compared to flexible tabular formats, the fixed format restricts table reasoning performance even with self-consistency.
% 并且，在使用Markdown时self-consistency的性能相比贪婪解码的性能提升不多甚至下降，这因为self-consistency在使用CoT推理时性能并不稳定，随着温度上升模型的指令遵循能力有所下降
Moreover, self-consistency with Markdown improves slightly or even decreases compared to greedy decoding.
This is attributed to self-consistency instability in CoT prompting \cite{chen2024selfconsistencyfailure,renze2024effectsamplingtemperatureproblem}, and diminished instruction following ability at higher sampling temperatures \cite{zeng2024evaluatinginstructionfollowing,peeperkorn2024temperaturecreativity}.
Additionally, the tables reveal that:

\paragraph{The improvement of \ourmethod on the more challenging dataset is more significant.}
% 我们的方法在WikiTQ上的提升比在TabFact上的提升显著
% 我们的方法在不同数据集上都带来了性能提升，证明了我们方法的有效性
Both \ourmethods and \ourmethodv achieve performance improvement across datasets, underscoring their efficacy.
% 此外，相对于较简单的TabFact数据集，Single方法在WikiTQ上的性能提升更显著
Specifically, the performance on WikiTQ is significantly better compared to the simpler TabFact.
% 因为较为简单的数据集中，被不同表示正确解决的instance不能很好地反映各个表示之间的区别，更具有随机性
% The limited improvement of \ourmethods on TabFact is attributed to the inability of this simpler dataset to highlight the differences between various formats, resulting in restricted classification performance. 
% 虽然在TabFact上分类的性能超过在WikiTQ上的性能，我们方法的提升较弱因为单一表示本身的性能较高，导致提升的空间较小
Despite higher classification accuracy on TabFact (Table~\ref{tab:classification}), \ourmethods shows limited improvement due to the already high-performance baseline. 
% The inability of simpler questions to highlight the differences between various formats also limits the classification performance.
% 并且，因为TabFact的问题更简单，导致被不同表格表示正确解决的instance之间的重叠更大（见附录），导致Vote方法在TabFact上的性能提升也更小
Moreover, \ourmethodv exhibits less improvement on TabFact due to the greater overlap between simper instances correctly solved by different formats (see supplementary material).

% 通用模型相较于代码模型的性能提升更显著
\paragraph{\ourmethods shows superior performance on the general model compared to the code model.}
% 我们的single和vote方法在不同模型上都带来了性能提升，证明了我们方法的有效性
\ourmethods and \ourmethodv demonstrate efficacy across diverse models.
% 此外，相对于code模型，在general模型上的性能提升更显著
Notably, \ourmethods exhibits greater improvement on the general models than on the code models.
% 因为code模型最适合的表格表示都是与代码相关的，区别相较于md这类自然语言的表格表示差别更小，因此分类器学习分类的效果更差
The tabular formats suitable for the code model are closely related to code, resulting in smaller differences and consequently limited classifier performance.

\paragraph{\ourmethodv consistently surpasses \ourmethods.}
% % 我们的Vote方法一致地超越了Single方法
% 虽然Single方法的性能受限于分类性能（见表），但有着较高的推理效率
Although the performance of \ourmethods is constrained by classification accuracy (see Table~\ref{tab:classification}), it demonstrates high reasoning efficiency which only infers once.
% Vote方法虽然推理性能较高，得益于多种表格表示的推理结果，但推理花销较大。
In contrast, \ourmethodv achieves superior table reasoning performance across various LLMs and datasets, because instances could be resolved correctly by multiple formats, and errors produced by different tabular formats exhibit diversity. 
% 但是Vote方法也因此推理效率较低
However, \ourmethodv is less efficient in reasoning.

\subsection{Analysis}
% 我们选取Llama3-8B进行后续分析实验，因为其较高的推理效率
We select Llama3-8B for subsequent analytical experiments due to its high reasoning efficiency. 
% 并且，分析实验在WikiTQ上进行，因为其问题更多样
Moreover, we conduct most experiments on WikiTQ, because it encompasses more diverse questions \citep{pasupat-liang-2015-WikiTableQuestions,2019TabFact,dong2022tablesurvey-plm,shi-etal-2020-squall}.

    % 统一示例
    % 表格对性能的影响是由于使用了不同的示例吗
    % \subsubsection{Unifying Demonstrations}
    \subsubsection{Is the impact of tabular formats due to the different demonstrations?}
        \label{subsub:unifying}

        \begin{table}[ht]
            \centering
            \small
            \begin{tabular}{cccccc}
    \toprule
    \textbf{MD} & \textbf{Dict} & \textbf{List} & \textbf{PD} & \textbf{DB} & \bm{$\text{Acc}_{\text{max}} - \text{Acc}_{\text{min}}$} \\
    \midrule
    $47.7$ & $43.0$ & $30.5$ & $39.2$ & $16.1$ & $31.6$ \\
    \bottomrule
\end{tabular}
            \caption{
            % 将表格表示成不同的形式，使用Llama3-8B在WikiTQ上的准确率
            The accuracy on the WikiTQ dataset using Llama3-8B with greedy decoding, employing unified demonstrations in the prompts, which are different from the prompts used in the main experiments. 
            MD denotes Markdown, PD denotes Pandas, and DB denotes Database. 
            % variance指的是相同模型下不同表格表示的性能最大差距
            % \textbf{Variance} refers to the maximum performance gap with different tabular formats.
            }
            \label{tab:unified demo}
        \end{table}
        
    % 为了证明我们的实验中表格表示对性能的影响不是因为示例，我们采用了统一的示例进行实验，实验结果如表所示
    % To determine whether the impact of tabular formats on experimental performance is independent of variations in demonstrations within prompts, we unify the demonstrations.
    We conduct experiments with unified demonstrations in the prompts for each format and present the results in Table~\ref{tab:unified demo}.
    % 具体地，我们在不同的表示对应的prompt中，采用相同问题的示例，并分别人工标注了对应的rationale和代码，具体的prompt在附录
    Specifically, we adopt the demonstrations of the same question across different tabular formats, manually annotating the rationale or program accordingly. 
    Detailed prompts are present in the supplementary material.
    % 可以发现，统一示例后，表格表示对性能的影响仍然很大，并且性能差距变得更大
    Table~\ref{tab:unified demo} indicates that the tabular formats continue to significantly affect performance, with the performance gap widening when using unified demonstrations.

    % 不同表格表示之间的重叠
    % 不同instance适合的表格表示是不同的吗
    % \subsubsection{Overlap between Tabular Formats}
    \subsubsection{Are the tabular formats suitable for different instances different?}
        % \begin{figure}[t]
        %     \centering
        %     \input{fig/representation_overlap}
        %     \caption{
        %     % 被各个表示正确解决的instance之间的重复率
        %     The overlap among instances that are correctly solved with each tabular format using Llama3-8B on WikiTQ.
        %     }
        %     \label{fig:representation_overlap}
        % \end{figure}

        \begin{figure}[t]
            \centering
            \includegraphics[width=0.6\linewidth]{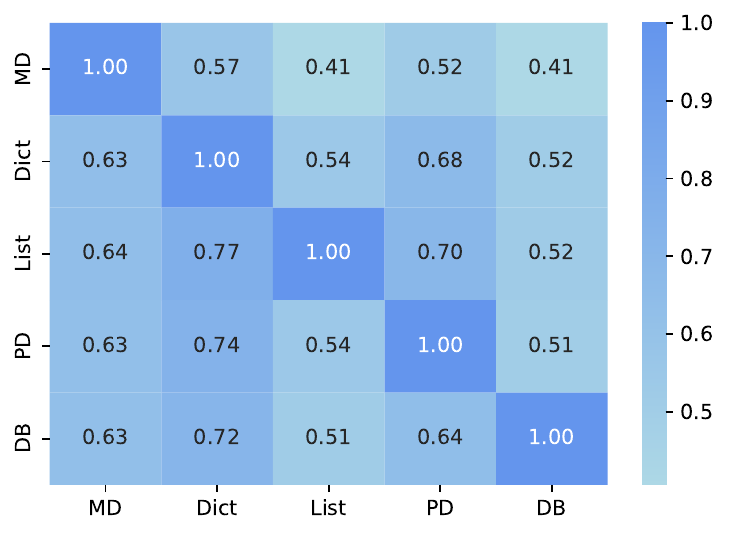}
            \caption{
            % 被各个表示正确解决的instance之间的重复率
            The overlap between instances solved by tabular formats achieved by Llama3-8B on WikiTQ. 
            % 图中的数值代表被横轴对应的表格正确解决的instance中，也能被纵轴对应的表格正确解决的比例
            The values represent the proportion of instances that can be solved by the tabular format corresponding to the vertical axis, within the instances solvable by the format on the horizontal axis.
            }
            \label{fig:representation_overlap}
        \end{figure}

        \begin{table}[t]
            \centering
            \small
            \begin{tabular}{ll|ccccc}
    \toprule
    \textbf{Model} & \textbf{Scale} & \textbf{MD} & \textbf{Dict} & \textbf{List} & \textbf{PD} & \textbf{DB} \\
    \midrule
    \multirow{2}{*}{Llama3} & 8B & $24.6$ & $7.1$ & $4.6$ & $5.9$ & $8.3$ \\
     & 70B & $16.0$ & $1.7$ & $2.2$ & $1.8$ & $4.3$ \\
    \midrule
    \multirow{2}{*}{DeepSeek-Coder} & 6.7B & $24.7$ & $6.7$ & $3.7$ &  $11.0$ & $13.6$ \\
     & 33B & $15.6$ & $3.9$ & $3.3$ & $4.4$ & $7.5$ \\
    \bottomrule
\end{tabular}
            \caption{
            % 能被每种表格表示正确解决的instance中，只能被这种表示正确解决的百分比
            The percentage of instances that can only be correctly solved by one tabular format, in all instances that the tabular format can correctly solve.
            }
            \label{tab:analysis_example}
        \end{table}

        % 为了探索我们方法有效的原因，我们分析了由不同表格表示正确解决的instance之间的重叠度，如图所示
        % To further claim that different instances suit different tabular formats, 
        We analyze the overlap between instances correctly solved by each format and the proportion of instances that can only be solved by a specific format, as shown in Figure~\ref{fig:representation_overlap} and Table~\ref{tab:analysis_example}. 
        % 我们发现
        Figure~\ref{fig:representation_overlap} illustrates that: 
        % 1. 由各个表示正确解决的instance与其他表示正确解决的instance之间的重叠度较小，均小于80%，说明各个表示之间的差异性较为明显，这进一步证明了我们的观点：适合求解不同instance的表格表示不同
        (\emph{i})~The overlaps between instances correctly solved by each format are all $\leq 80\%$, indicating the distinctions among formats.
        % , which further claims that different formats are suitable for different instances. 
        % 因此，结合观察表1的oracle一列，为每个instance选择合适的表格表示相比直接选择该数据集上性能最好的表示，性能上升的空间更大
        % Therefore, selecting the suitable tabular format for each instance offers greater potential for performance improvement compared to using a fixed format for all instances.
        % 2. 与其他表格表示对应的instance重合度较高的是JSON和PD，因为这两个表示对应的表格推理性能较高；而重合度较低的是DB和DM，对应的性能也较低
        (\emph{ii})~The Dict and Pandas formats exhibit higher overlap due to their superior performance (see Table~\ref{tab:main}), while the DB and List formats have lower overlap. 
        The overlaps between instances solved by different formats using four LLMs on WikiTQ and TabFact are provided in the supplementary material.

        Table~\ref{tab:analysis_example} presents that: 
        % 1. 存在一定比例的instance只能被某种表示正确解决，使用不同的模型，进一步证明了不同表格表示之间存在的差异
        (\emph{i})~Certain instances can only be accurately solved using a specific format, highlighting the differences between formats.
        % 2. 只能被MD和DB正确解决的比例普遍比Dict，List和PD的比例大，因为MD和DB这两种表示和另外3种表示的差异较大
        % (\emph{ii})~Markdown and Database formats uniquely solve the higher proportion of instances compared to the Dict, List, and Pandas formats due to their distinct structure from these programming formats. 
        (\emph{ii})~Markdown and Database formats resolve a greater proportion of instances compared to Dict, List, and Pandas formats due to their distinct structures compared to the programming formats. 
        % 图4和表5说明了我们的观点，不同的instance适合不同的表格表示求解
        Figure~\ref{fig:representation_overlap} and Table~\ref{tab:analysis_example} claim that different instances suit different tabular formats.

    % 候选表格数量怎么影响我们方法的性能
    % \subsubsection{Number of Candidate Tabular Formats}
    \subsubsection{How does the number of candidate tabular formats affect \ourmethod?}
        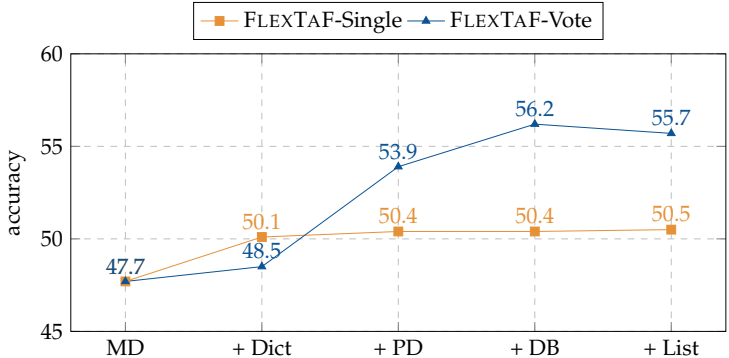
\begin{figure}
            \centering          \resizebox{0.7\linewidth}{!}{
\begin{tikzpicture}
        \begin{axis}[
            width=12cm,
            height=6cm,
            ymin=45, ymax=60,
            xtick=data,
            xticklabels={MD, + Dict, + PD, + DB, + List},
            ytick={45, 50, 55, 60},
            bar width=20pt,
            ylabel={accuracy},
            nodes near coords,
            legend style={at={(0.5,1.05)},anchor=south,legend columns=-1},
            grid=both, % 添加此行以启用网格线
            grid style=dashed % 可选：设置网格线的样式为虚线
        ]
        % First line plot data
        \addplot[color=single, mark=square*] coordinates {
            (1, 47.7)
            (2, 50.1)
            (3, 50.4)
            (4, 50.4)
            (5, 50.5)
        };

        % Second line plot data
        \addplot[color=vote, mark=triangle*] coordinates {
            (1, 47.7)
            (2, 48.5)
            (3, 53.9)
            (4, 56.2)
            (5, 55.7)
        };
        % Third line plot data
        % \addplot[color=corange, mark=triangle*] coordinates {
        %     (1, 47.7)
        %     (2, 63.5)
        %     (3, 68.2)
        %     (4, 71.2)
        %     (5, 72.6)
        % };
        % Bar chart data
        % \addplot[ybar, fill=cpurple] coordinates {
        %     (1, 47.7)
        %     (2, 43.0)
        %     (3, 39.2)
        %     (4, 31.0)
        %     (5, 30.5)
        % };
        \legend{\ourmethods, \ourmethodv}
        \end{axis}
    \end{tikzpicture}
}

% \resizebox{1\linewidth}{!}{
% \begin{tikzpicture}
%         \begin{axis}[
%             width=12cm,
%             height=8cm,
%             ymin=30, ymax=55,
%             xtick=data,
%             xticklabels={MD, Dict, PD, Tuple, DB, List},
%             ytick={30, 35, 40, 45, 50, 55},
%             bar width=20pt,
%             ylabel={accuracy},
%             nodes near coords,
%             legend style={at={(0.5,1.05)},anchor=south,legend columns=-1}
%         ]
%         % First line plot data
%         \addplot[color=cgreen, mark=square*] coordinates {
%             (1, 47.7)
%             (2, 50.9)
%             (3, 50.4)
%             (4, 50.8)
%             (5, 50.6)
%             (6, 50.2)
%         };

%         % Second line plot data
%         \addplot[color=cblue, mark=triangle*] coordinates {
%             (1, 47.7)
%             (2, 47.7)
%             (3, 53.9)
%             (4, 49.8)
%             (5, 53.0)
%             (6, 53.6)
%         };
%         % Bar chart data
%         \addplot[ybar, fill=cpurple] coordinates {
%             (1, 47.7)
%             (2, 43.0)
%             (3, 39.2)
%             (4, 33.6)
%             (5, 31.0)
%             (6, 30.5)
%         };
%         \legend{\ourmethods, \ourmethodv}
%         \end{axis}
%     \end{tikzpicture}
% }
            \caption{
            % 使用Llama3-8B在WikiTQ上单一表示的性能
            % The accuracy using Llama3-8B with a single tabular format on WikiTQ.
            % 随着从左至右的每个表示加入候选，我们方法的性能
            The accuracy of \ourmethods and \ourmethodv using Llama3-8B on WikiTQ with different numbers of candidate tabular formats, as additional candidate tabular formats are added from left to right.
            }
            \label{fig:representation_number}
        \end{figure}
        % 为了探索候选表格表示数量对我们方法的影响，我们在不同数量的候选representations的条件下进行实验，结果如图所示
        % To explore how the number of candidate tabular formats influences \ourmethod, 
        We perform experiments using varying numbers of candidate formats and present results in Figure~\ref{fig:representation_number}, where formats are added in descending order of performance.
        % 我们按照性能由高到低的顺序依次加入候选
        % 受限于计算资源，我们只在最多5种表示上进行实验，并且我们方法的性能已经趋近平缓。
        % 我们发现
        We observe the following: 
        % 1. 我们的Single方法随着候选表格数量的变化，性能逐渐上升至平稳，因为不同的表格表示都有只能被这种表示解决的instance，所以
        % 随着表格表示数量的增加，Oracle的性能逐渐增加，待分类的数据逐渐增加，但分类的性能并没有一直上升，分类的难度更大
        (\emph{i})~The performance of \ourmethods gradually stabilizes as the number of candidate formats increases. 
        The classification performance does not always improve due to the increased difficulty with a higher number of labels \cite{large-scale-multi-label,long-tailed-multi-label}.
        % 2. 我们的Vote方法随着表格数量的增加，性能波动较大，因为其性能与每种表示的性能较为相关
        (\emph{ii})~\ourmethodv varies greatly with the increase in the number of formats due to its reliance on the performance of each format.
        % 尤其是，当Markdown和Dict两种表示作为候选的时候，Vote方法没有超过Single方法，因为只有两种表示时Vote方法面对不同的结果只能通过prob选择一种表示，不能很好反应答案的正误
        In particular, \ourmethodv does not exceed \ourmethods when Markdown and Dict are candidates, as it selects from two different answers only by comparing probabilities, which do not accurately indicate correctness \cite{wang2023selfconsistency,estimating-confidence,quevedo2024detectinghallucinations_probability}. 
        % 综合考虑Single和Vote方法，我们主实验选取了5种表示作为候选format。
        Therefore, we select $5$ formats as candidates in the main experiments, considering the performance of our two methods.

    % 每个表格表示的分类性能
    % 如何进一步提升我们方法的分类性能
    % \subsubsection{Performance of Classification}
    \subsubsection{How to further improve the classification performance of \ourmethods?}
        \begin{table*}[t]
            \small
            \centering
            \begin{tabular}{l|cc|cc|cc|cc}
    \toprule
    \multirow{3}{*}{\textbf{Format}} & \multicolumn{4}{c|}{\textbf{WikiTQ}} & \multicolumn{4}{c}{\textbf{TabFact}} \\
    & \multicolumn{2}{c|}{\textbf{Llama3}} & \multicolumn{2}{c|}{\textbf{DeepSeek-Coder}} & \multicolumn{2}{c|}{\textbf{Llama3}} & \multicolumn{2}{c}{\textbf{DeepSeek-Coder}} \\
    & 8B & 70B & 6.7B & 33B & 8B & 70B & 6.7B & 33B \\
    \midrule
    Markdown & \bm{$78.7$} & \bm{$73.8$} & $47.1$ & $20.3$ & \bm{$64.3$} & \bm{$67.1$} & $25.3$ & \bm{$36.0$} \\
    Dict & $27.9$ & $23.8$ & $35.1$ & \bm{$46.7$} & $25.0$ & $14.3$ & \bm{$37.5$} & \bm{$36.0$}\\
    List &  $3.4$ & $29.6$ & $9.7$ & $19.4$ & $20.0$ & $25.0$ & $0.0$ & $0.0$\\
    Pandas & $2.1$ & $10.0$ & $28.6$ & $30.1$ & $0.0$ & $11.1$ & $22.7$ & $7.7$ \\
    Database & $14.3$ & $15.6$ & \bm{$50.6$} & $29.7$ & $18.5$ & $26.7$ & $31.3$ & $29.6$\\
    \midrule
    % Overall & $69.6 (2193)$ & $82.2 (2983)$ & $68.5 (2013)$ & $73.1 (2369)$ & $79.6 (1539)$ & $88.8 (1740)$ & $74.3 (1417)$ & $80.6 (1565)$\\
    Overall & $69.6$ & $82.2$ & $68.5$ & $73.1$ & $79.6$ & $88.8$ & $74.3$ & $80.6$\\
    \bottomrule
\end{tabular}

% \begin{tabular}{lll|ccccc|c}
%     \toprule
%     \textbf{Dataset} & \textbf{Model} & \textbf{Scale} & \textbf{MD} & \textbf{Dict} & \textbf{List} & \textbf{PD} & \textbf{DB} & \textbf{Avg.} \\
%     \midrule
%     \multirow{4}{*}{WikiTQ} & \multirow{2}{*}{Llama3} & 8B &  &  &  &  &  &  \\
%     & & 70B &  &  &  &  &  &  \\
%     \cmidrule{2-9}
%     & \multirow{2}{*}{DeepSeek-Coder} & 6.7B &  &  &  &  &  & \\
%     & & 33B &  &  &  &  &  & \\
%     \midrule
%     \multirow{4}{*}{TabFact} & \multirow{2}{*}{Llama3} & 8B &  &  &  &  &  & \\
%     & & 70B &  &  &  &  &  & \\
%     \cmidrule{2-9}
%     & \multirow{2}{*}{DeepSeek-Coder} & 6.7B &  &  &  &  &  & \\
%     & & 33B &  &  &  &  &  & \\        
%     \bottomrule
% \end{tabular}
            \caption{
            % Single方法在不同表示上的准确率，以及平均准确率
            The overall classification accuracy of \ourmethods using four LLMs on two datasets, and the classification accuracy for instances that can only be correctly solved by a single format, with the best performance marked in \textbf{bold}. 
            }
            \label{tab:classification}
        \end{table*}
        % 为了讨论如何进一步提升我们方法的分类性能，我们统计了整体分类的性能和标签数量为1的instance的性能，因为这一部分数据最能反映该种表示的特征，如表所示
        % To discuss how to further improve the classification performance of \ourmethods, 
        We analyze both the overall accuracy and the accuracy on instances that can be correctly solved by only one format, as these instances most distinctly highlight the unique features of each format.
        The classification results are presented in Table~\ref{tab:classification}. 
        % 我们发现
        We observe that:
        % 预测较大规模的LLM时的分类性能优于较小规模的LLM，这是因为较大规模的LLM有着较好的鲁棒性，导致用相同format正确解决的instance有着更一致的特征和更少的随机性
        (\emph{i})~The classification performance of predicting larger-scale LLMs is better than that of smaller-scale LLMs. 
        This is attributed to the greater robustness of larger-scale LLMs \cite{howe2024robustness_llm}, which results in more consistent features in instances correctly solved with the same format.
        % 2. 分类器预测各个表示的准确率与各个表示的性能呈正相关，因为各个标签的训练数据的size和该表示的性能呈正相关
        (\emph{ii})~The classification performance of each format is positively correlated with the size of its training data (see the supplementary material). 
        % 所以，未来可以通过减少噪声数据和扩大训练数据规模来提升分类器的性能，以及在各个标签上的性能
        Therefore, future improvements in classification performance can be achieved by reducing noise in the data and increasing the scale of the training data.

        % Our observations are as follows: (\emph{i})~Larger-scale LLMs demonstrate better classification performance than smaller-scale LLMs, likely due to the greater robustness of the former \cite{howe2024robustness_llm}, which leads to more consistent features in instances correctly solved by a single format. (\emph{ii})~The classification accuracy of each format is positively correlated with the size of its training data (see the supplementary material). Thus, future enhancements in classification performance could be achieved by reducing data noise and expanding the training data scale.

    % 训练集标签的数量
    % 训练集标签的数量怎样影响我们方法的性能
    % \subsubsection{Number of Classification Label}
    \subsubsection{How does the data filtering strategy affect \ourmethods?}
        \begin{figure}
            \centering
            % \small
            \resizebox{0.75\linewidth}{!}{
\begin{tikzpicture}
\begin{axis}[
    axis y line*=left,
    ymin=59, ymax=71,
    ytick={59, 61, 63, 65, 67, 69, 71},
    ylabel={Accuracy of Classification},
    axis x line*=bottom,
    xlabel={Maximum threshold of the number of labels},
    xtick={1, 2, 3, 4, 5},
    xmin=0.5, xmax=5.5,
    xmajorgrids,
    ymajorgrids,
    grid style=dashed,
    width=12cm,
    height=8cm,
    legend style={
        at={(0.5,1.1)},
        anchor=south,
        legend columns=-1,
        /tikz/every even column/.append style={column sep=1cm}
    }
]

\addplot[color=blue, mark=*] 
    coordinates {(1,65.7) (2,68.5) (3,70.0) (4,69.2) (5,59.9)};
\addlegendentry{Accuracy of Classification}

\addplot[color=red, mark=square*] 
    coordinates {(1,65.7)};
\addlegendentry{Accuracy of \ourmethods}

\end{axis}

\begin{axis}[
    axis y line*=right,
    ymin=41, ymax=53,
    ytick={41, 43, 45, 47, 49, 51, 53},
    ylabel={Accuracy of \ourmethods},
    axis x line=none,
    width=12cm,
    height=8cm,
    xmin=0.5, xmax=5.5,
]

\addplot[color=red, mark=square*] 
    coordinates {(1,47.7) (2,49.7) (3,50.5) (4,50.3) (5,43.5)};
% \addlegendentry{Accuracy of \ourmethods}

\end{axis}

\end{tikzpicture}
}
            \caption{
            % 随着训练集中标签数量上限的变化，分类的准确率和我们方法的准确率
            The accuracy of classification and \ourmethods, with the change of the maximum threshold of the number of labels in the training data. 
            }
            \label{fig:number of label}
        \end{figure}
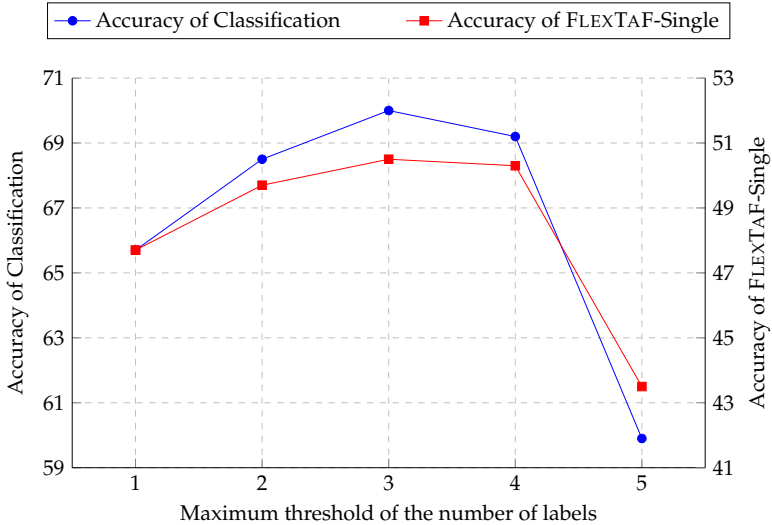  
    % 为了探索采用不同的训练数据的过滤策略对性能的影响，我们分别采取了不同的过滤方式，即变化训练数据的标签的数量的上限，实验结果如图所示
    We compare various strategies by adjusting the maximum threshold of the number of labels in each training data instance.
    The experimental results are shown in Figure~\ref{fig:number of label}.
    % 可以发现，
    We find that: 
    % % 1. 随着训练数据中标签数量上限的增长，整体的性能首先呈现上升的趋势，直至上限为3，证明了选用最多3种标签进行训练的有效性。
    % (\emph{i})~With the increase of the maximum threshold of the number of labels, overall performance first improves until the threshold is set to $3$, which suggests the effectiveness of selecting data with $\leq3$ labels for training.
    % % 因为随着标签数量上限的上升，更多的训练数据有利于模型的训练
    % This is because more training data is conducive to the training process.
    % % This suggests that selecting data with $\leq3$ labels for training is effective, as more training data facilitates the learning process.
    % % 2. 当标签上限大于3时，性能逐渐下降，证明了训练数据中采取过滤措施的必要性。
    % (\emph{ii})~When the maximum threshold exceeds $3$, the performance gradually declines, underscoring the importance of the data filtering strategy.
    (\emph{i})~As the maximum label count rises, overall performance improves, peaking at $3$ labels, which demonstrates that training with $\leq3$ labels per instance works best, because the increased amount of training data aids the model training.
    (\emph{ii})~When the label count goes above $3$, performance drops. 
    % 因为当某个instance能被大多数表示解决时，这个instance往往较为简单，无法很好地体现各个标签的特征，影响模型的训练
    % As instances solvable by most tabular formats tend to be simpler and fail to capture the distinct characteristics of each label, these instances introduce noise into the training process.
    This shows why filtering training data is important, as instances easily resolved by most tabular formats are often simpler and fail to effectively capture the unique features of each label, thus negatively impacting model training.

\subsection{Case Study}
    \begin{figure}[t]
        \centering
        \includegraphics[width=0.5\linewidth]{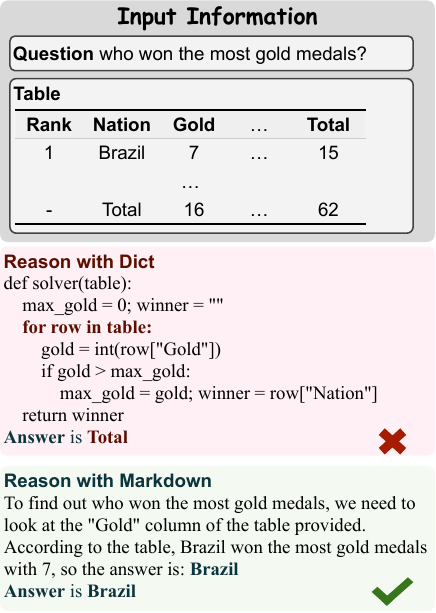}
        \caption{
        An instance from the WikiTQ test set using Llama3-70B with Markdown and Dict tabular formats.
        }
        \label{fig:case}
    \end{figure}
    % 为了更直观地理解不同表格推理的instance适合的表格表示不同，我们展示了一个WikiTQ的例子
    To better illustrate that different instances are suitable for different tabular formats, we present an instance from WikiTQ. 
    % 在使用Dict表示表格时，Llama3-70B产生了错误的求解代码，which遍历了表格的所有行，没有排除特殊的“-”一行
    As illustrated in Figure~\ref{fig:case}, when utilizing the Dict format, Llama3-70B~\cite{Llama3} generates an incorrect program that processes all table rows without excluding a special "-" row.
    % 而在用Markdown求解时，模型则正确地找到了获得金牌最多的国家，成功地排除了“-”一行的干扰
    Conversely, when solving the instance with Markdown, the model successfully ignores the "-" line and correctly identifies the country that won the most gold medals.
    % 所以，在使用Llama3-70B求解这个问题时，Markdown比Dict更适合表示这个表格
    % Therefore, for this instance, Markdown is a more suitable format than Dict when using Llama3-70B.
    Therefore, for this instance, the Markdown format proves more suitable than Dict when employing Llama3-70B.
    % 这也反映了我们的观点，表格在instance粒度上对表格推理的性能有影响
    % This demonstrates that different instances are suitable for different tabular formats. 
    Additional instances are provided in the supplementary material.

    \section{Related Works}
        \label{sec:related}
            % 表格推理任务指根据表格完成用户的自然语言需求
    The table reasoning task aims to answer the natural language question based on provided tabular data \citep{dong2022tablesurvey-plm,zhang2024surveytablereasoning,dong2024tablesurvey}.
    % 目前，基于LLM的方法由于其优越的性能已经成为表格推理任务的主流方法
    LLM-based methods have become predominant in this field due to their superior performance of table understanding, commonsense reasoning, and logical reasoning \citep{chain-ot-thought,chen-2023-few-one,zhao-etal-2023-parseres,pot,pal}.
    % 有的研究者们关注于通过微调LLM以增强模型的表格推理能力
    Some existing works focus on enhancing table reasoning ability by fine-tuning LLMs \citep{zhang2024tablellama,bian2024hellama,TableLLM,patnaik2024cabinet,ProTrix,gardner2024largescaletransferlearning,li2024table-gpt}.
    % 而有的研究者通过设计prompt，分解任务或集合多个推理结果，以提升表格推理性能
    Also, some works improve table reasoning performance by designing prompts \citep{zhao-etal-2023-parseres,dater,nahid2024normtab,nahid2024tabsqlify,lee2024reduce_table,Wang2024chainoftable,zhao-etal-2024-tapera} or aggregating diverse results \citep{ni2023lever,liu-etal-2024-rethinking}.
    % 比如，Mix self-consistency发现用自然语言推理和用程序语言推理适用于求解不同的问题，因此提出集合两种推理的结果，以进一步提升模型表格推理的性能
    For instance, Mix Self-Consistency~\citep{liu-etal-2024-rethinking} proposes combining textual and symbolic reasoning results.
    % to enhance table reasoning.

    % 总之，前人工作在训练或推理时，往往无论使用哪个模型，面对不同的问题和表格，都固定一种表格表示
    Previous works generally employ a fixed tabular format across varying instances, regardless of the specific LLM used, which could limit the performance. 
    % 他们发现不同的表格表示在不同的表格理解任务，比如Cell Lookup，Size Detection上有着不同的性能，并提出了self-augmented prompting
    Since \citet{TableMeetsLLM} demonstrate that the performance of table understanding tasks, such as Cell Lookup and Size Detection, varies with different tabular formats, and propose self-augmented prompting.
    % 为了帮助LLM理解表格，self-augmented prompting使用LLM抽取表格中重要的内容，并用LLM将其总结成自然语言描述
    To aid LLMs in understanding tables, self-augmented prompting employs LLMs to extract critical table contents and summarize them into natural language descriptions for LLMs.
    % 前人在此基础上探索了更多表格表示在表格理解任务上的性能，以及施加不同的噪声操作，比如ShuffleColumns，TransposeTable的影响
    Building on this, \citet{singha2023tabular-representation} evaluate the performance of additional tabular formats and investigate the effects of noise operations like Shuffle Columns and Transpose Table on table understanding. 
    % 他们发现模型使用不同的表格，在不同任务上的鲁棒性也不同
    They analyze that the LLM adopts different tabular formats, causing varying robustness in table understanding tasks. 
    Similarly, \citet{deng-etal-2024-tables-texts-images} examine the different performance of text-based and image-based formats in table reasoning tasks.

    % \citet{TableMeetsLLM} demonstrate that the effectiveness of table understanding tasks, such as Cell Lookup and Size Detection, varies with different tabular formats. They propose self-augmented prompting, which enhances table reasoning by extracting key table contents and summarizing them into natural language for LLMs. Building on this, \citet{singha2023tabular-representation} evaluate the performance of additional tabular formats and investigate the effects of noise operations like Shuffle Columns and Transpose Table on table understanding. They find that different tabular formats lead to varying robustness in LLM performance. Similarly, \citet{deng-etal-2024-tables-texts-images} examine the differential performance of text-based and image-based formats in table reasoning tasks.

    % Since \citet{TableMeetsLLM,singha2023tabular-representation} suggest that different table reasoning tasks benefit from distinct tabular formats. 
    % 比如，他们使用GPT3进行实验，发现了使用不同表格表示在不同任务上的性能不同
    % For instance, \citet{singha2023tabular-representation} utilize GPT-3~\cite{gpt3} in their experiments and demonstrate that performance varies with different tabular formats depending on the task.

    % 然而，他们只从任务的角度探讨了不同表格表示的影响，忽略了具体的instance和使用的模型对表格表示性能的影响
    However, existing studies primarily discuss the impact of different tabular formats from the perspective of tasks, ignoring the impact of the instances and LLMs.
    % 所以我们提出我们的方法，通过变化表格的表示来提升模型表格推理的性能
    Therefore, we claim that different instances and models suit different tabular formats by analyzing the experimental results.
    Based on the claim, we propose \ourmethod to improve the table reasoning performance by flexibly employing tabular formats.

    \section{Conclusion}
    % 在本文，我们从两方面探索了表格表示对表格推理性能的影响
    In this paper, we explore the impact of tabular format on table reasoning performance from two aspects.
    % 1. 我们从实验结果的角度证明了使用不同LLM在instance粒度上有自己合适的表格表示
    (\emph{i})~We claim from the perspective of experimental results that different instances and models have different most suitable tabular formats. 
    % 2. 我们提出了我们的方法
    (\emph{ii})~We propose our methods.
    % Single方法通过训练的分类器预测不同LLM推理适合求解每个instance的表格表示
    \ourmethods use the classifier to predict the most suitable tabular format according to the instance and the LLM.
    % Vote方法对多个表示得到的结果进行最大投票得到最终结果
    \ourmethodv obtains the results from multiple formats and employs the voting mechanism to get the final result.
    % 我们在WikiTQ和TabFact数据集上构建实验
    We build experiments on WikiTableQuestions and TabFact datasets.
    % 相比使用单一表示的最好性能，我们的方法分别平均提升%和%，证明了我们方法的有效性
    Compared with the best performance of using a fixed tabular format with greedy decoding and self-consistency, \ourmethods and \ourmethodv increase by $2.3\%$ and $4.8\%$ on average respectively, demonstrating the effectiveness of our methods.
    
    \clearpage

    \bibliography{colm2024_conference}
    \bibliographystyle{colm2024_conference}
    \clearpage

    \appendix
    % 五种表格表示的介绍
\section{Introduction to Five Tabular Formats}
    \label{app:tabular_formats}
    % 在本节，我们介绍我们方法中采用的5种表格表示：Markdown, Dict, List, Pandas, and Database
    In this section, we introduce the five tabular formats used in \ourmethod: Markdown, Dict, List, Pandas, and Database.
    % Markdown格式是指用md语言表示表格
    The Markdown format refers to the representing tables in the Markdown language.
    % Dict格式采用List[Dict: [str, Any]]索引表格，其中每一行用Dict存储，每一列的列名作为Dict的key被索引
    The Dict format employs \texttt{List[Dict: [str, Any]]} to index the table, in which each row is stored in a dictionary, and the column name of each column is indexed as the key of the dictionary.
    % List格式格式采用List[List[Any]]索引表格，其中每一行用，包括表头用List存储，每一列需要用列的序号被索引
    The List format adopts \texttt{List[List[Any]]} to index the table, in which each row, including the header, is stored in the list, and each column needs to be indexed with the serial number of the column.
    % Pandas格式是用Pandas DataFrame API定义表格的Python代码片段，其中指出了表格的每一行，以及表头
    The Pandas format is a Python code snippet that uses the Pandas DataFrame API to define the table, which points out each line of the table and the header.
    % Database格式是指将表格表示成数据库的格式，用CREATE语句描述了列名，并列举具体的value
    Database format refers to the format of representing a table as a database, describing the column name with a CREATE statement, and listing specific values.

\section{Prompts with Each Tabular Format}
We present the prompts used in experiments in this section.
    \label{app:prompts}
    \subsection{Prompts for Main Experiments}
    
    \begin{table*}[t]
        \centering
        \small
        \begin{tabular}{p{0.9\textwidth}}
            \toprule
            \textbf{The prompt with the Markdown format for WikiTQ.} \\
            \midrule
            Please answer the question with the given table, present the final result in the format "..., so the answer is: (answer)":\\
            Please note that utilize the format, do not include periods. Here are some instances you may refer to:\\
            ---\\
            table:\\
            $\mid Aircraft \mid Description \mid Max Gross Weight \mid Total disk area \mid Max disk Loading \mid$\\
            $\mid:---\mid:---\mid:---\mid:---\mid:---\mid$\\
            $\mid Robinson R-22 \mid Light utility helicopter \mid 1,370 lb (635 kg) \mid ... \mid$\\
            $\mid Bell 206B3 JetRanger \mid Turboshaft utility helicopter \mid 3,200 lb (1,451 kg) \mid ... \mid$\\
            $\mid CH-47D Chinook \mid Tandem rotor helicopter \mid 50,000 lb (22,680 kg) \mid ... \mid$\\
            utterance:\\
            What is the max gross weight of the Robinson R-22?\\
            answer:\\
            To find out what is the max gross weight of Robinson R-22, we need to look at the "Max Gross Weight" column of the table provided. According to the table, the max gross weight of the Robinson R-22 is 1,370 lb (635 kg), so the answer is: 1,370 lb (635 kg)\\
            ---\\
            table:\\
            $\mid Player \mid No. \mid Nationality \mid Position \mid Years in Toronto \mid School/Club Team \mid$\\
            $\mid:---\mid:---\mid:---\mid:---\mid:---\mid:---\mid$\\
            $\mid Mark Baker \mid 3 \mid United States \mid Guard \mid 1998-99 \mid Ohio State \mid$\\
            $\mid Marcus Banks \mid 3 \mid United States \mid Guard \mid 2009-10 \mid UNLV \mid$\\
            $\mid Leandro Barbosa \mid 20 \mid Brazil \mid Guard \mid 2010-2012 \mid Tilibra/Copimax ( Brazil ) \mid$\\
            utterance:\\
            How many players were with the school or club team La Salle?\\
            answer: \\
            To count the number of players with the school or club team La Salle, we need to look at the "School/Club Team" column of the table provided. 
            According to the table, there are 1 player whose School/Club Team is La Salle, so the answer is: 1\\
            ---\\
            table:\\
            $\mid Model \mid 1991 \mid 1995 \mid 1996 \mid 1997 \mid 1998 \mid 1999 \mid 2000 \mid 2001 \mid 2002 \mid 2003 \mid 2004 \mid$\\
            $\mid:---\mid:---\mid:---\mid:---\mid:---\mid:---\mid:---\mid:---\mid:---\mid:---\mid:---\mid:---\mid$\\
            $\mid Skoda Felicia \mid 172,000 \mid 210,000 \mid - \mid 288,458 \mid 261,127 \mid 241,256 \mid 148,028 \mid 44,963 \mid - \mid$\\
            $\mid Skoda Octavia \mid - \mid - \mid - \mid 47,876 \mid 102,373 \mid 143,251 \mid 158,503 \mid 164,134 \mid 164,017 \mid 165,635 \mid 181,683 \mid$\\
            $\mid Skoda Fabia \mid - \mid - \mid - \mid - \mid - \mid 823 \mid 128,872 \mid 250,978 \mid 264,641 \mid 260,988 \mid 247,600 \mid$\\
            is the number on skoda fabia for 1999 more or less than 1000?\\
            answer:\\
            To find out if the number on the Skoda Fabia for 1999 is more or less than 1000, we need to look at the data provided for the "Skoda Fabia" in "1999", which is 823, that is, the number for the Skoda Fabia in 1999 is less than 1000, so the answer is: less\\
            ---\\
            table:\\
            $\mid Place \mid Rider \mid Country \mid Team \mid Points \mid Wins \mid$\\
            $\mid:---\mid:---\mid:---\mid:---\mid:---\mid:---\mid$\\
            $\mid 1 \mid Sylvain Geboers \mid Belgium \mid Suzuki \mid 3066 \mid 3 \mid$\\
            $\mid 2 \mid Adolf Weil \mid Germany \mid Maico \mid 2331 \mid 2 \mid$\\
            $\mid 3 \mid John Banks \mid United Kingdom \mid CZ \mid 971 \mid 0 \mid$\\
            $\mid 4 \mid Mark Blackwell \mid United States \mid Husqvarna \mid 604 \mid 0 \mid$\\
            $\mid 5 \mid Brad Lackey \mid United States \mid CZ \mid 603 \mid 0 \mid$\\
            utterance:\\
            which country had the most riders that placed in the top 20 of the 1971 trans-ama final standings?\\
            answer:\\
            To find out which country had the most riders in the top 20, we need to look at the "Country" column of the table provided and count the number of times each country appears. 
            According to the table, United States had the most riders, so the answer is: United States\\
            ---\\
            table:\\
            $<$table$>$\\
            utterance:\\
            $<$utterance$>$\\
            answer:\\
            \bottomrule
        \end{tabular}
        \caption{The prompt with the Markdown format used in the main experiments for WikiTQ.}
        \label{tab:prompt_md}
    \end{table*}

    \begin{table*}[t]
        \centering
        \small
        \begin{tabular}{p{0.9\textwidth}}
            \toprule
            \textbf{The prompt with the Dict format for WikiTQ.} \\
            \midrule
            Answer the question with the given table using python code.\\
            You should generate a function with the following signature without any other parameters. Here are some instances you may refer to:\\
            ---\\
            \begin{verbatim}
table = [
    {
        "Aircraft": "Robinson R-22",
        "Max Gross Weight": "1,370 lb (635 kg)",
        ...
    },
    ...
]
utterance: What is the max gross weight of the Robinson R-22?
def solver(table):
    for row in table:
        if row["Aircraft"] == "Robinson R-22":
            return row["Max Gross Weight"]
            \end{verbatim}
            ---
            \begin{verbatim}
table = [...]
utterance: How many players were with the school or club team La Salle?
def solver(table):
    players_la_salle = set()
    for row in table:
        if row["School/Club Team"] == "La Salle": players_la_salle.add(row["Player"])
    return len(players_la_salle)
            \end{verbatim}
            ---
            \begin{verbatim}
table = [...]
utterance: is the number on skoda fabia for 1999 more or less than 1000?
def solver(table):
    for row in table:
        if row["Model"] == "Skoda Fabia": num_1999 = row["1999"].replace(",", "")
            if int(num_1999) > 1000: return "more"
            else: return "less"
    return "less"  
            \end{verbatim}
            ---
            \begin{verbatim}
table = [...]
utterance: which country had the most riders that placed in the top 20 of the 1971 
trans-ama final standings?
def solver(table):
    country_counts = {}
    for row in table:
        country = row["Country"]
        if country in country_counts: country_counts[country] += 1
        else: country_counts[country] = 1
    max_riders = max(country_counts.values())
    countries_with_max_riders = [country for country, count in country_counts.items() 
    if count == max_riders]
    return countries_with_max_riders[0]
            \end{verbatim}
            ---\\
            Based on the above demonstrations, answer the following utterance with the following table using Python code.\\
            table = $<$table$>$\\
            utterance: $<$utterance$>$\\
            def solver(table):\\
                \# Your code here  \\
            \bottomrule
        \end{tabular}
        \caption{
        The prompt with the Dict format used in the main experiments for WikiTQ.
        % 受限于文章长度，我们没有列出示例中全部表格的内容
        Due to the limited length of the paper, we do not list the contents of all the tables in the demonstrations, which are the same tables in Table~\ref{tab:prompt_md}. 
        % 我们会在后续放出完整prompt
        }
        \label{tab:prompt_dict}
    \end{table*}
    
    \begin{table*}[t]
        \centering
        \small
        \begin{tabular}{p{0.9\textwidth}}
            \toprule
            \textbf{The prompt with the List format for WikiTQ.} \\
            \midrule
            Answer the question with the given table using python code.\\
            You should generate a function with the following signature without any other parameters. Here are some instances you may refer to:\\
            ---\\
            \begin{verbatim}
table = [
    [
        "Aircraft",
        ...
    ],
    [
        "Robinson R-22",
        ...
    ],
    ...
]
utterance: What is the max gross weight of the Robinson R-22?
def solver(table):
    for row in table[1:]:
        if row[0] == "Robinson R-22": return row[2]
            \end{verbatim}
            ---
            \begin{verbatim}
table = [...]
utterance: How many players were with the school or club team La Salle?
def solver(table):
    players_la_salle = set()
    for row in table[1:]:
        if row[5] == "La Salle": players_la_salle.add(row[0])
    return len(players_la_salle)
            \end{verbatim}
            ---
            \begin{verbatim}
table = [...]
utterance: is the number on skoda fabia for 1999 more or less than 1000?
def solver(table):
    for row in table[1:]:
        if row[0] == 'Skoda Fabia':
            if row[6] == "-" or int(row[6].replace(",", "")) < 1000: return "less"
            else:mreturn "more"
            \end{verbatim}
            ---
            \begin{verbatim}
table = [...]
utterance: which country had the most riders that placed in the top 20 of the 1971 
trans-ama final standings?
def solver(table):
    country_counts = {}
    for row in table[1:]:
        country = row[2]
        if country in country_counts: country_counts[country] += 1
        else: country_counts[country] = 1
    max_riders = max(country_counts.values())
    countries_max = [c for c, count in country_counts.items() if count == max_riders]
    return countries_max[0]
            \end{verbatim}
            ---\\
            Based on the above demonstrations, answer the following utterance with the following table using Python code.\\
            table = $<$table$>$\\
            utterance: $<$utterance$>$\\
            def solver(table):\\
                \# Your code here  \\
            \bottomrule
        \end{tabular}
        \caption{
        The prompt with the List format used in the main experiments for WikiTQ.
        % 受限于文章长度，我们没有列出示例中全部表格的内容
        Due to the limited length of the paper, we do not list the contents of all the tables in the demonstrations, which are the same tables in Table~\ref{tab:prompt_md}. 
        % 我们会在后续放出完整prompt
        }
        \label{tab:prompt_list}
    \end{table*}
    
    \begin{table*}[t]
        \centering
        \small
        \begin{tabular}{p{0.9\textwidth}}
            \toprule
            \textbf{The prompt with the Pandas format for WikiTQ.} \\
            \midrule
            Answer the question with the given table using python code.\\
            You should generate a function with the following signature without any other parameters. Here are some instances you may refer to:\\
            ---\\
            \begin{verbatim}
table = pd.DataFrame([
    [
        "Robinson R-22",
        ...
    ],
    ...
], columns = [
    "Aircraft",
    ...
]
)
utterance: What is the max gross weight of the Robinson R-22?
def solver(table):
    import pandas as pd
    max_gross_weight_r22 = table[table["Aircraft"] == "Robinson R-22"]
    ["Max Gross Weight"].iloc[0]
    return max_gross_weight_r22
            \end{verbatim}
            ---
            \begin{verbatim}
table = [...]
utterance: How many players were with the school or club team La Salle?
def solver(table):
    import pandas as pd
    la_salle_count = table[table["School/Club Team"] == "La Salle"].shape[0]
    return la_salle_count
            \end{verbatim}
            ---
            \begin{verbatim}
table = [...]
utterance: is the number on skoda fabia for 1999 more or less than 1000?
def solver(table):
    import pandas as pd
    fabia_row = table[table['Model'] == 'Skoda Fabia']
    fabia_1999_sales = fabia_row['1999'].values[0]
    if fabia_1999_sales == '-' or int(fabia_1999_sales.replace(",", "")) < 1000:
        return 'less'
    else:
        return 'more'
            \end{verbatim}
            ---
            \begin{verbatim}
table = [...]
utterance: which country had the most riders that placed in the top 20 of the 1971 
trans-ama final standings?
def solver(table):
    import pandas as pd
    most_riders_country = table['Country'].value_counts().idxmax()
    return most_riders_country
            \end{verbatim}
            ---\\
            Based on the above demonstrations, answer the following utterance with the following table using Python code.\\
            table = $<$table$>$\\
            utterance: $<$utterance$>$\\
            def solver(table):\\
                \# Your code here  \\
            \bottomrule
        \end{tabular}
        \caption{
        The prompt with the Pandas format used in the main experiments for WikiTQ.
        % 受限于文章长度，我们没有列出示例中全部表格的内容
        Due to the limited length of the paper, we do not list the contents of all the tables in the demonstrations, which are the same tables in Table~\ref{tab:prompt_md}. 
        % 我们会在后续放出完整prompt 
        }
        \label{tab:prompt_pd}
    \end{table*}

    \begin{table*}[t]
        \centering
        \small
        \begin{tabular}{p{0.9\textwidth}}
            \toprule
            \textbf{The prompt with the Database format for WikiTQ.} \\
            \midrule
            Please complete the sql below to solve the question with the given database.\\
            Here are some instances you may refer to:\\
            ---\\
            \begin{verbatim}
database:
CREATE TABLE information (
year int ,
division int ,
...
);
/*
Columns and instances in each column :
year: 1998, 1999, 2000, 2001, 2002, 2003, 2004, 2005, 2006, 2007, 2008, ... ;
...
*/
utterance:
when was the first time the kansas city brass qualified for the playoffs? 
sql:
SELECT year FROM information WHERE playoffs != 'Did not qualify' ORDER 
BY year ASC LIMIT 1;
            \end{verbatim}
            ---
            \begin{verbatim}
database:
CREATE TABLE information (...);
/*...*/
utterance:
what was the next episode after \"do-si-do?\"  
sql: 
SELECT episode FROM information WHERE num = (SELECT num FROM information 
WHERE episode = 'Do-Si-Do') + 1;
\end{verbatim}
---
\begin{verbatim}
database:
CREATE TABLE information (...);
/*...*/   
utterance:
which dino album yielded the most songs on the billboard hot 100?      
sql:
SELECT album FROM information WHERE chart = 'Billboard Hot 100' GROUP BY album 
ORDER BY COUNT(*) DESC LIMIT 1;
            \end{verbatim}
            ---
            \begin{verbatim}
database:
CREATE TABLE information (...);
/*...*/
utterance:
when was the last year team penske finished first?
sql:
SELECT MAX(year) FROM information WHERE team = 'Team Penske' AND finish = 1;
            \end{verbatim}
            ---\\
            Based on the above demonstrations, answer the following utterance with the following database using SQL.\\
            database:\\
            $<$table$>$\\
            utterance:\\
            $<$utterance$>$\\
            sql:\\
            SELECT\\
            \bottomrule
        \end{tabular}
        \caption{
        The prompt with the Database format used in the main experiments for WikiTQ.
        % 受限于文章长度，我们没有列出示例中全部表格的内容
        Due to the limited length of the paper, we do not list the contents of all the tables in the demonstrations. 
        % 我们会在后续放出完整prompt       
        }
        \label{tab:prompt_db}
    \end{table*}

    \begin{table*}[t]
        \centering
        \small
        \begin{tabular}{p{0.9\textwidth}}
            \toprule
            \textbf{The prompt with the Markdown format on TabFact.} \\
            \midrule
            Verify the consistency between the table and the utterance.\\
            Please present the final result in the format "..., so the answer is: (answer)" and the "(answer)" is "True" or "False".\\
            Please note that utilize the format, do not include periods.\\
            Here are some demonstrations you may refer to:\\
            ---\\
            table:\\
            $\mid tournament \mid wins \mid top - 5 \mid top - 10 \mid top - 25 \mid events \mid cuts made \mid$\\
            $\mid:---\mid:---\mid:---\mid:---\mid:---\mid:---\mid:---\mid$\\
            $\mid masters tournament \mid 0 \mid 1 \mid 2 \mid 4 \mid 4 \mid 4 \mid$\\
            $\mid us open \mid 0 \mid 2 \mid 3 \mid 4 \mid 6 \mid 5 \mid$\\
            $\mid the open championship \mid 1 \mid 2 \mid 2 \mid 2 \mid 3 \mid 3 \mid$\\
            $\mid pga championship \mid 0 \mid 0 \mid 1 \mid 2 \mid 5 \mid 4 \mid$\\
            $\mid totals \mid 1 \mid 5 \mid 8 \mid 12 \mid 18 \mid 16 \mid$\\            
            utterance: \\
            tony lema be in the top 5 for the master tournament , the us open , and the open championship\\
            answer:\\
            To verify whether tony lema be in the top 5 for the master tournament , the us open , and the open championship, we need to look at the "top - 5" column of the table provided. According to the table, the "top - 5" column of the "masters tournament", "us open", and "the open championship" are all more than zero, so the answer is: True\\
            ---\\
            table:\\
            $\mid year \mid competition \mid venue \mid position \mid event \mid$\\
            $\mid:---\mid:---\mid:---\mid:---\mid:---\mid$\\
            $\mid 2006 \mid world cross country championships \mid fukuoka , japan \mid 10th \mid individual junior race \mid$\\
            $\mid 2006 \mid world cross country championships \mid fukuoka , japan \mid 3rd \mid team junior race \mid$\\
            $\mid 2006 \mid african championships in athletics \mid bambous , mauritius \mid 5th \mid 10000 m \mid$\\
            $\mid 2006 \mid world road running championships \mid debrecen , hungary \mid 7th \mid individual 20 km \mid$\\
            $\mid 2006 \mid world road running championships \mid debrecen , hungary \mid 3rd \mid team 20 km \mid$\\
            $\mid 2007 \mid world cross country championships \mid mombasa , kenya \mid 7th \mid individual \mid$\\
            $\mid 2007 \mid all - africa games \mid algiers , algeria \mid 2nd \mid 10000 m \mid$\\
            $\mid 2007 \mid world championships in athletics \mid osaka , japan \mid 13th \mid 10000 m \mid$\\
            $\mid 2009 \mid world cross country championships \mid amman , jordan \mid 17th \mid individual \mid$\\
            $\mid 2013 \mid world championships \mid moscow , russia \mid 3rd \mid marathon \mid$\\
            utterance:\\
            japan and hungary host the competition 3 time each\\    
            answer:\\
            To verify whether japan and hungary both host the competition 3 time, we need to look at the "venue" column of the table provided. According to the table, "japan" hosts the competition 3 times, but "hungary" hosts the competition 2 times, so the answer is: False\\
            ---\\
            Based on the above demonstrations, Verify the consistency between the following table and utterance.\\
            table:\\
            $<$table$>$\\
            utterance:\\
            $<$utterance$>$\\
            answer:\\
            \bottomrule
        \end{tabular}
        \caption{The prompt with the Markdown format used in the main experiments for TabFact.}
        \label{tab:prompt_md_tabfact}
    \end{table*}

    \begin{table*}[t]
        \centering
        \small
        \begin{tabular}{p{0.9\textwidth}}
            \toprule
            \textbf{The prompt with the Dict format for TabFact.} \\
            \midrule
            Verify the consistency between the table and the utterance with "True" or "False" using python code.\\
            You should generate a function with the following signature without any other parameters:\\
            Here are some demonstrations you may refer to:\\
            ---\\
            \begin{verbatim}
table = [
    {
        "tournament": "masters tournament",
        "wins": "0",
        "top - 5": "1",
        "top - 10": "2",
        "top - 25": "4"
    },
    {
        "tournament": "us open",
        "wins": "0",
        "top - 5": "2",
        "top - 10": "3",
        "top - 25": "4"
    },
    ...
]
utterance: tony lema be in the top 5 for the master tournament , the us open , 
and the open championship
def solver(table):
    top_5_tournament = [row["tournament"] for row in table if int(row["top - 5"]) > 0]
    if "masters tournament" not in top_5_tournament:
        return False
    if "us open" not in top_5_tournament:
        return False
    if "the open championship" not in top_5_tournament:
        return False
    return True
            \end{verbatim}
            ---
            \begin{verbatim}
table = [
    {
        "year": "2006",
        "competition": "world cross country championships",
        "venue": "fukuoka , japan",
        "position": "10th",
        "event": "individual junior race"
    },
    ...
]
utterance: japan and hungary host the competition 3 time each
def solver(table):
    japan_host_time = 0
    hungary_host_time = 0
    for row in table:
        if "japan" in row["venue"]:
            japan_host_time += 1
        elif "hungary" in row["venue"]:
            hungary_host_time += 1
    return (japan_host_time == 3 and hungary_host_time == 3)
            \end{verbatim}
            ---\\
            Based on the above demonstrations, Verify the consistency between the following table and utterance.\\
            table = $<$table$>$\\
            utterance: $<$utterance$>$\\
            def solver(table):\\
                \# Your code here  \\
            \bottomrule
        \end{tabular}
        \caption{
        The prompt with the Dict format used in the main experiments for TabFact.
        }
        \label{tab:prompt_dict_tabfact}
    \end{table*}

    \begin{table*}[t]
        \centering
        \small
        \begin{tabular}{p{0.9\textwidth}}
            \toprule
            \textbf{The prompt with the List format for TabFact.} \\
            \midrule
            Verify the consistency between the table and the utterance with "True" or "False" using python code.\\
            You should generate a function with the following signature without any other parameters:\\
            Here are some demonstrations you may refer to:\\
            ---\\
            \begin{verbatim}
table = [
    [
        "tournament",
        "wins",
        "top - 5",
        "top - 10",
        "top - 25"
    ],
    [
        "masters tournament",
        "0",
        "1",
        "2",
        "4"
    ],
    ...
]
utterance: tony lema be in the top 5 for the master tournament , the us open , 
and the open championship
def solver(table):
    top_5_tournament = [row[0] for row in table[1:] if int(row[2]) > 0]
    if "masters tournament" not in top_5_tournament:
        return False
    if "us open" not in top_5_tournament:
        return False
    if "the open championship" not in top_5_tournament:
        return False
    return True
            \end{verbatim}
            ---
            \begin{verbatim}
table = [
    [
        "year",
        "competition",
        "venue",
        "position",
        "event"
    ],
    ...
]
utterance: japan and hungary host the competition 3 time each
def solver(table):
    japan_host_time = 0
    hungary_host_time = 0
    for row in table[1:]:
        if "japan" in row[2]:
            japan_host_time += 1
        elif "hungary" in row[2]:
            hungary_host_time += 1
    return (japan_host_time == 3 and hungary_host_time == 3)
            \end{verbatim}
            ---\\
            Based on the above demonstrations, answer the following utterance with the following table using Python code.\\
            table = $<$table$>$\\
            utterance: $<$utterance$>$\\
            def solver(table):\\
                \# Your code here  \\
            \bottomrule
        \end{tabular}
        \caption{
        The prompt with the List format used in the main experiments for TabFact.
        }
        \label{tab:prompt_list_tabfact}
    \end{table*}

    \begin{table*}[t]
        \centering
        \small
        \begin{tabular}{p{0.9\textwidth}}
            \toprule
            \textbf{The prompt with the Pandas format for TabFact.} \\
            \midrule
            Verify the consistency between the table and the utterance with "True" or "False" using python code.\\
            You should generate a function with the following signature without any other parameters:\\
            Here are some demonstrations you may refer to, and you don't need to answer the demonstrations:\\
            ---\\
            \begin{verbatim}
table = pd.DataFrame([
    [
        "masters tournament",
        "0",
        "1",
        "2",
        "4",
        "4",
        "4"
    ],
    ...
], columns = [
    "tournament",
    "wins",
    "top - 5",
    "top - 10",
    "top - 25",
    "events",
    "cuts made"
]
)
utterance: tony lema be in the top 5 for the master tournament , the us open , 
and the open championship
def solver(table):
    tournaments = ["masters tournament", "us open", "the open championship"]
    for tournament in tournaments:
        row = table[table['tournament'] == tournament]
        if row.empty or int(row['top - 5'].values[0]) == 0:
            return False 
    return True
            \end{verbatim}
            ---\\
            Based on the above demonstrations, answer the following utterance with the following table using Python code.\\
            table = $<$table$>$\\
            utterance: $<$utterance$>$\\
            def solver(table):\\
                \# Your code here  \\
            \bottomrule
        \end{tabular}
        \caption{
        The prompt with the Pandas format used in the main experiments for TabFact.
        }
        \label{tab:prompt_pd_tabfact}
    \end{table*}

    \begin{table*}[t]
        \centering
        \small
        \begin{tabular}{p{0.9\textwidth}}
            \toprule
            \textbf{The prompt with the Database format for TabFact.} \\
            \midrule
            Please complete the sql below to solve the question with the given database.\\
            Here are some instances you may refer to:\\
            ---\\
            \begin{verbatim}
CREATE TABLE information (
tournament text ,
wins int ,
top_5 int ,
top_10 int ,
top_25 int ,
events int ,
cuts_made int
);
/*
Columns and instances in each column :
tournament: masters tournament, us open, the open championship, pga championship, totals ;
wins: 0, 0, 1, 0, 1 ;
top_5: 1, 2, 2, 0, 5 ;
top_10: 2, 3, 2, 1, 8 ;
top_25: 4, 4, 2, 2, 12 ;
*/

utterance: 
tony lema be in the top 5 for the master tournament , the us open , 
and the open championship

SQL:
SELECT CASE  WHEN (SELECT COUNT(*) FROM information WHERE tournament IN 
('masters tournament', 'us open', 'the open championship') AND top_5 > 0) = 3 
THEN 'True' ELSE 'False' END AS result;
            \end{verbatim}
            ---
            \begin{verbatim}
CREATE TABLE information (
year int ,
competition text ,
venue text ,
position text ,
event text
);
/*
Columns and instances in each column :
year: 2006, 2006, 2006, 2006, 2006, 2007, 2007, 2007, 2009, 2013 ;
competition: world cross country championships, world cross country championships, ... ;
venue: fukuoka , japan, fukuoka , japan, bambous , mauritius, debrecen , hungary, ... ;
position: 10th, 3rd, 5th, 7th, 3rd, 7th, 2nd, 13th, 17th, 3rd ;
event: individual junior race, team junior race, 10000 m, ... ;
*/

utterance: 
japan and hungary host the competition 3 time each

SQL:
SELECT CASE WHEN (SELECT COUNT(*) FROM information WHERE venue LIKE '%japan%') = 3 AND 
(SELECT COUNT(*) FROM information WHERE venue LIKE '%hungary%') = 3 THEN 'True' 
ELSE 'False' END AS result;
\end{verbatim}
            ---\\
            Based on the above demonstrations, answer the following utterance with the following database using SQL.\\
            database:\\
            $<$table$>$\\
            utterance:\\
            $<$utterance$>$\\
            sql:\\
            SELECT\\
            \bottomrule
        \end{tabular}
        \caption{
        The prompt with the Database format used in the main experiments for TabFact.
        }
        \label{tab:prompt_db_tabfact}
    \end{table*}

    The prompts for WikiTQ~\cite{pasupat-liang-2015-WikiTableQuestions} in main experiments with a single tabular format are shown in Table~\ref{tab:prompt_md}, Table~\ref{tab:prompt_dict}, Table~\ref{tab:prompt_list}, Table~\ref{tab:prompt_pd}, and Table~\ref{tab:prompt_db}. 
    And the prompts for TabFact~\cite{2019TabFact} in main experiments with a single tabular format are shown in Table~\ref{tab:prompt_md_tabfact}, Table~\ref{tab:prompt_dict_tabfact}, Table~\ref{tab:prompt_list_tabfact}, Table~\ref{tab:prompt_pd_tabfact}, and Table~\ref{tab:prompt_db_tabfact}. 
    % 可以发现，只有用数据库表示表格的时候用的示例不一样，因为如果用和其他表示相同的示例，用于求解的SQL会很复杂，降低模型的推理性能
    It can be found that only the demonstrations with the Database format are different among the prompts for WikiTQ.
    If we use the same demonstrations as other prompts, the SQL given in the prompt is too complicated to reduce the reasoning performance of the model.
    
    \subsection{Prompts for Experiments of Unifying Demonstrations}

    \begin{table*}[t]
        \centering
        \small
        \begin{tabular}{p{0.9\textwidth}}
            \toprule
            \textbf{The prompt with the Database format using unified demonstrations.} \\
            \midrule
            Please complete the sql below to solve the question with the given database.\\
            Here are some instances you may refer to:\\
            ---\\
            \begin{verbatim}
database:
CREATE TABLE information (
aircraft text ,
...
);
/*
Columns and instances in each column :
aircraft: Robinson R-22, Bell 206B3 JetRanger, CH-47D Chinook, ... ;
...
*/
utterance:
What is the max gross weight of the Robinson R-22?
sql:
SELECT max_gross_weight FROM information WHERE aircraft = 'Robinson R-22';
            \end{verbatim}
            ---
            \begin{verbatim}
database:
CREATE TABLE information (
player text ,
...
);
/*...*/
utterance:
How many players were with the school or club team La Salle?         
sql: 
SELECT COUNT(*) FROM information WHERE school_club_team = 'La Salle';
            \end{verbatim}
            ---
            \begin{verbatim}
database:
...
/*...*/   
utterance:
is the number on skoda fabia for 1999 more or less than 1000?
sql:
SELECT CASE WHEN CAST(column_1999 AS INTEGER) < 1000 THEN 'less' ELSE 'more' 
END AS result FROM information WHERE model = 'Skoda Fabia';
            \end{verbatim}
            ---
            \begin{verbatim}
database:
...
utterance:
which country had the most riders that placed in the top 20?
sql:
SELECT country, COUNT(rider) as rider_count FROM information WHERE place <= 20
GROUP BY country ORDER BY rider_count DESC LIMIT 1;
            \end{verbatim}
            ---\\
            Based on the above demonstrations, answer the following utterance with the following database using SQL.\\
            database:\\
            $<$table$>$\\
            utterance:\\
            $<$utterance$>$\\
            sql:\\
            SELECT\\
            \bottomrule
        \end{tabular}
        \caption{
        The prompt with the Database format with the unified demonstrations.
        % 受限于文章长度，我们没有列出示例中全部表格的内容
        Due to the limited length of the paper, we do not list the contents of all the tables in the demonstrations, which are the same tables in Table~\ref{tab:prompt_md}. 
        % 我们会在后续放出完整prompt
        % 
        }
        \label{tab:prompt_db_unified}
    \end{table*}
    
    % 我们在本节展示统一示例后的prompt
    We present the prompt with the unified demonstrations in this subsection.
    % 从表6789可知，MD, Dict, List, 和PD用的示例是统一的，所以我们只改变了数据库的prompt
    From Table~\ref{tab:prompt_md}, Table~\ref{tab:prompt_dict}, Table~\ref{tab:prompt_list} and Table~\ref{tab:prompt_pd}, it can be seen that the demonstrations used in the Markdown, Dict, List, and Pandas are unified, so we only change the prompt with the Database format, which is shown in Table~\ref{tab:prompt_db_unified}.

% 表格representation在模型、instance上不同分布的证明
\section{Tabular Format Distributions on Models}
    \label{app:proof_analysis}
% 在本节，我们展示在第2节中证明表格representation在模型、instance上分布不同的具体过程
In this section, we show the specific process of proving that the tabular format is distributed differently on the model, which is discussed in \S2. 

% 具体地，我们把每个模型上正确表示的数量作为观测频率，计算每个表示在不同模型下平均正确的数量作为预期概率，并计算卡方统计量
Specifically, we take the number correctly represented on each model as the observed frequency $O_i$, calculate the average correct number of each format under different models as the expected frequency $E_i$, and calculate the Chi-square statistics as Equation~\ref{eqa:x2}.
\begin{equation}
    \mathcal{X}^2 = \sum \frac{(O_i-E_i)^2}{E_i}
    \label{eqa:x2}
\end{equation}
% 其对应的自由度则为
The corresponding degree of freedom $dof$ is $dof = (N_m-1) * (N_r-1)$, where $N_m$ is the number of models, and $N_r$ is the number of tabular formats.
% 根据计算出的卡方统计量和对应的自由度，从卡方分布表中查找对应的 p 值，which小于0.05，证明在不同模型上的正确表格表示的分布存在差异
According to the calculated Chi-square $\mathcal{X}^2$ and the corresponding degree of freedom $dof$, we find the corresponding P-value from the square distribution table, which is less than $0.05$, proving that there is a distinct difference in the distribution of the correct tabular formats on different models.

% \begin{table}[t]
%     \centering
%     \begin{tabular}{@{}llcc@{}}
%     \toprule
%      & & \textbf{WikiTQ} & \textbf{TabFact} \\ \midrule
%     \multicolumn{2}{l}{\textbf{Pre-Training}} & & \\
%      & batch size & $32$ & $128$ \\
%      & max epoch & $1$ & $1$ \\
%      & learning rate & $1e-6$ & $1e-6$ \\
%      & weight decay & $0.01$ & $0.01$ \\
%      & max tokens & $512$ & $512$ \\
%      & warm up steps & $500$ & $500$ \\
%     \midrule
%     \multicolumn{2}{l}{\textbf{Training}} & & \\
%      & batch size & $128$ & $128$ \\
%      & epoch & $200$ & $400$ \\
%      & learning rate & $1e-5$ & $5e-6$ \\
%      & weight decay & $0$ & $0$ \\
%      & max tokens & $512$ & $512$ \\
%      & warm up steps & $0$ & $0$ \\ \bottomrule
%     \end{tabular}
%      \caption{
%      Training Parameters for training on WikiTQ and TabFact.
%      }
%      \label{tab:training_detail}
% \end{table}

\begin{table*}[t]
    \centering
    \begin{tabular}{@{}ll|cc@{}}
    \toprule
     & & \textbf{WikiTQ} & \textbf{TabFact} \\
    \midrule
    % \multicolumn{2}{l}{\textbf{Training}} & & \\
     & batch size & $128$ & $128$ \\
     & epoch & $200$ & $400$ \\
     & learning rate & $1e-5$ & $5e-6$ \\
     & max tokens & $512$ & $512$ \\
     & training device & $2 \times$ NVIDIA A100 40G GPU & $2 \times$ NVIDIA A100 40G GPU \\
     & training time & $16h$ & $36h$ \\ 
     \bottomrule
    \end{tabular}
     \caption{
     Classification training details in \ourmethods on WikiTQ and TabFact.
     }
     \label{tab:training_detail}
\end{table*}

\begin{table*}[h!]
  \centering
  \begin{tabular}{c|c|c|c|c|c|c|c|c}
    \toprule
    & \multicolumn{4}{c|}{\textbf{WikiTQ}} & \multicolumn{4}{c}{\textbf{TabFact}} \\
    & \multicolumn{2}{c|}{\textbf{Llama3}} & \multicolumn{2}{c|}{\textbf{DeepSeek-Coder}} & \multicolumn{2}{c|}{\textbf{Llama3}} & \multicolumn{2}{c}{\textbf{DeepSeek-Coder}} \\
    & 8B & 70B & 6.7B & 33B & 8B & 70B & 6.7B & 33B \\
    \midrule
    Data Size & $7,247$ & $5,649$ & $7,848$ & $5,944$ & $48,073$ & $26,050$ & $45,194$ & $29,619$ \\
    \bottomrule
  \end{tabular}
  \caption{
  The training data size employed for training the classifier in \ourmethods.
  }
  \label{tab:data size}
\end{table*}

% 表格分类器的训练细节
\section{Details of Training the Classifier in \ourmethods}

\begin{figure*}[t]
    \centering
    \includegraphics[width=1\linewidth]{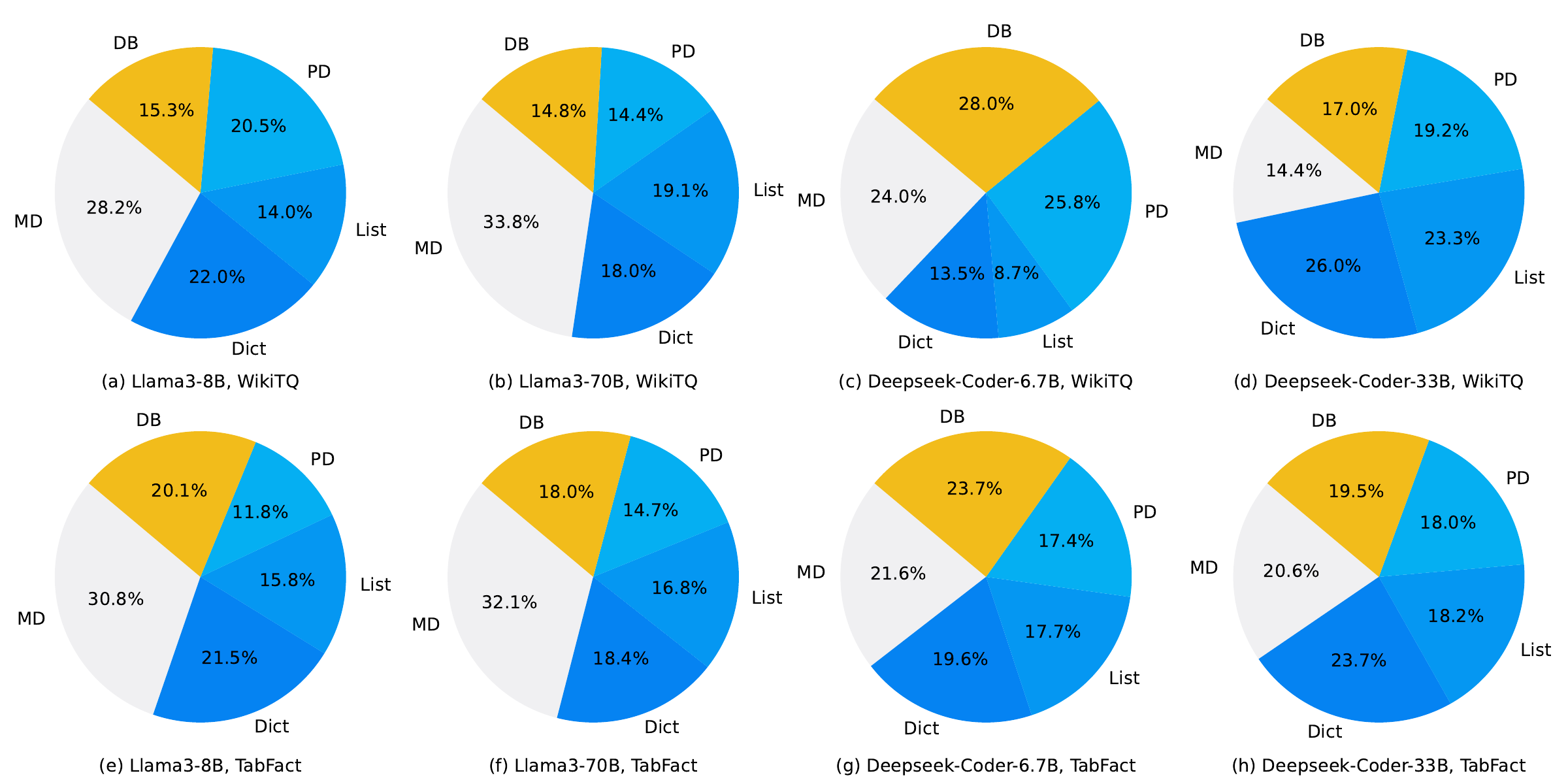}
    \caption{
    % 训练集中各个表示的比例
    The proportion of each tabular format in the training data.
    }
    \label{fig:proportion-plots}
\end{figure*}
    \label{app:training details}
    % 在本节，我们介绍我们方法中训练分类器的细节，如表所示
    In this section, we introduce the details of training the classifier in \ourmethods. 
    % With the annotated data used for training, the classifier could struggle to predict the most suitable tabular format limited by the scale of the training data.
    % % 受限于训练数据的规模，为了增加分类器的性能，我们为其增加了预训练任务：预测表格的表示，给定由某种表示表示的表格，模型输出这种表示的名字
    % To aid the classifier in understanding the tabular formats, we introduce a pre-training task: predicting the format according to the table, which needs to predict the name of this format given a table represented by a certain format.
    % % 具体地，我们随机生成了300,000个由随机字符串组成的表格，将表格用5种表示分别表示，再从所有数据中随机选择100,000个表格作为预训练的训练数据
    % Specifically, we randomly generate $300,000$ tables composed of random strings, represent the tables separately in the $5$ formats respectively, and then randomly select $100,000$ instances from all the data as the pre-training data. 
    % 我们统计了用于预测各个LLM在不同数据集上的训练数据的size在表中
    Our model is implemented with PyTorch~\cite{PyTorch} and transformers~\cite{wolf-etal-2020-transformers}. 
    We present training details in Table~\ref{tab:training_detail}, and we summarize the training data size employed for training the classifier to predict each LLM across various datasets in Table~\ref{tab:data size}. 
    % 我们统计了在不同数据集上预测不同LLM的训练数据中，各个表格表示的比例，如图所示
    We count the proportion of each tabular format in the training data for different LLMs and datasets, as shown in Figure~\ref{fig:proportion-plots}.
    % 具体地，为了便于比较不同表示的比例，我们统计每个表示占训练数据中所有示例的所有标签的比例
    Specifically, to compare the proportions of different tabular formats in the training data, we calculate the proportion of each format to all labels of all instances in the training data.

\section{Overlap between Tabular Formats}
    \label{app:oracle}
        \begin{figure*}[ht]
            \centering
            \includegraphics[width=1\linewidth]{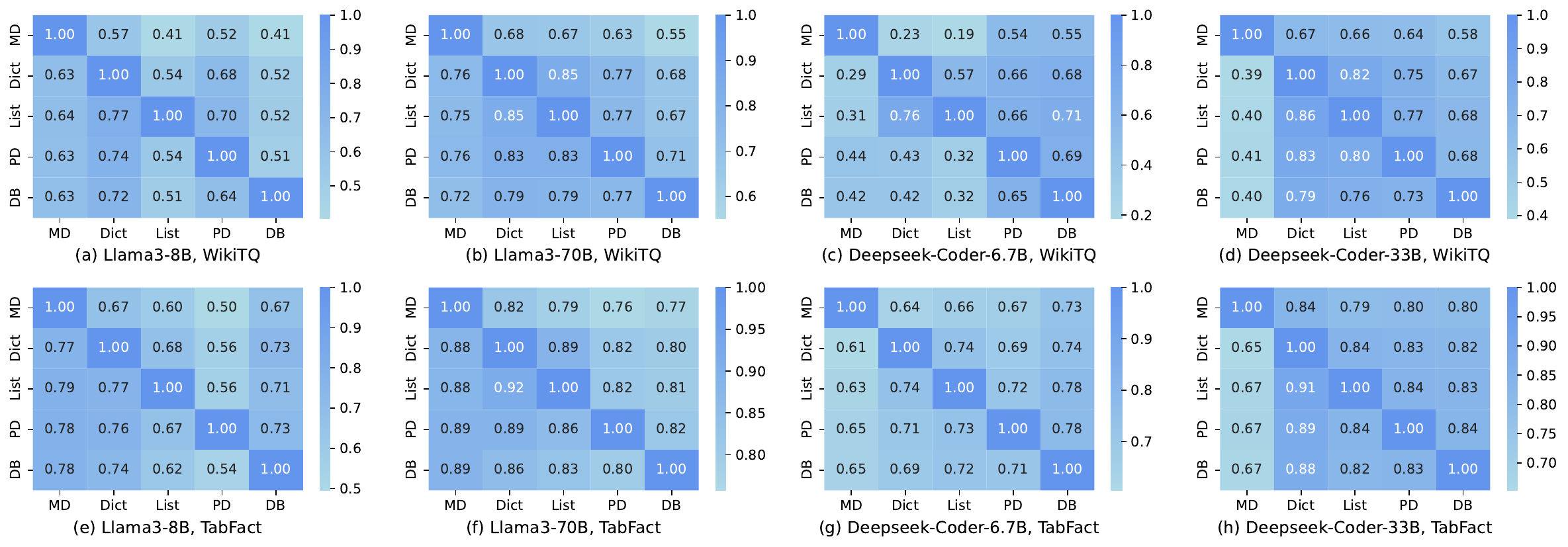}
            \caption{
            % 被各个表示正确解决的instance之间的重复率
            The overlap among instances that are correctly solved with each tabular format using four LLMs on two datasets. 
            % 图中的数值代表被横轴对应的表格正确解决的instance中，也能被纵轴对应的表格正确解决的比例
            The value represents the proportion that can be solved by the tabular format corresponding to the vertical axis in the instances that are solved by the format corresponding to the horizontal axis.
            }
            \label{fig:representation_overlap_all}
        \end{figure*}
    % 在本节，我们分析了the overlap between instances correctly solved by different tabular formats
    In this section, we analyze the overlap between instances correctly solved by different tabular formats, which is shown in Figure~\ref{fig:representation_overlap_all}. 
    % 由不同表格表示解决的instance虽然存在重叠，但均小于100%，证明了不同的instance适合不同的表格表示
    The overlap between instances solved by different tabular formats is all $\leq 100\%$, which proves that different instances are suitable for different tabular formats. 
    We can find that: 
    % 1. 使用DeepSeek-Coder造成的重叠比Llama3的更大，因为DeepSeek-Coder更擅长使用代码格式，代码格式如Dict,List之间的差别比代码格式和自然语言格式，如MD的差别更小
    (\emph{i})~The overlap caused by using DeepSeek-Coder is greater than that caused by Llama3 because DeepSeek-Coder is better at code formats. 
    The difference between code formats such as Dict and List is smaller than that between code formats and natural language formats, such as Markdown.
    % 2. 使用较大规模的LLM造成的重叠比较小规模的LLM更严重，因为较大规模的LLM表格推理性能越高，被不同表格表示正确解决的instance也越多
    (\emph{ii})~The overlap caused by employing large-scale LLMs is more serious than that of small-scale LLMs because the large-scale LLMs have higher table reasoning performance, leading to more instances correctly solved by tabular formats.
    % 3. 在TabFact上不同表格表示的重叠比在WikiTQ上不同表格表示的重叠大，因为TabFact上的问题更简单，容易被更多表格表示解决
    (\emph{iii})~The overlap on TabFact is greater than that on WikiTQ because the questions of TabFact are simpler and easier to solve by more tabular formats.

\section{Comparison \ourmethod with Oracle}
    \label{app:oracle}

    \begin{table*}[t]
        \centering
        \small
        \begin{tabular}{l|cc|cc|cc|cc}
            \toprule
            \multirow{3}{*}{\textbf{Tabular Format}} & \multicolumn{4}{c|}{\textbf{WikiTQ}} & \multicolumn{4}{c}{\textbf{TabFact}} \\
            & \multicolumn{2}{c|}{\textbf{Llama3}} & \multicolumn{2}{c|}{\textbf{DeepSeek-Coder}} & \multicolumn{2}{c|}{\textbf{Llama3}} & \multicolumn{2}{c}{\textbf{DeepSeek-Coder}} \\
            & 8B & 70B & 6.7B & 33B & 8B & 70B & 6.7B & 33B \\
            \midrule
            \ourmethods & $50.5$ & $69.1$ & $46.3$ & $54.5$ & $77.0$ & $87.1$ & $70.9$ & $78.3$ \\
            \ourmethodv & \bm{$55.7$} & \bm{$69.9$} & \bm{$51.4$} & \bm{$60.9$} & \bm{$80.3$} & \bm{$88.5$} & \bm{$77.9$} & \bm{$84.4$} \\
            Oracle & $71.8$ & $83.5$ & $67.7$ & $74.6$ & $96.9$ & $98.1$ & $95.4$ & $97.1$ \\
            \bottomrule
        \end{tabular}
        \caption{
        The performance of \ourmethods, \ourmethodv, and Oracle, which denotes the result that each instance uses the tabular format that can get the correct answer. 
        }
        \label{tab:oracle}
    \end{table*}

    % 在本节，我们把我们方法的性能和Oracle方法的性能比较，如表所示
    In this section, we compare the performance of \ourmethod with that of Oracle, as shown in Table~\ref{tab:oracle}.
    % 可以发现
    We observe that:
    % 1. Oracle的优秀性能说明了不同表示之间的差异较大，被不同表格表示正确解决的instance之间的overlap较小，进一步说明了不同的instance都有各自适合的表格表示
    (\emph{i})~The excellent oracle performance shows that there exist differences between different formats, which further suggests that different instances have their suitable tabular formats.
    % 2. 我们的Single方法受限于分类性能，与Oracle性能差距较大
    (\emph{ii})~The performance of \ourmethods is limited by classification and has a gap with the oracle performance, since the Pre-trained Language Model (PLM) cannot predict the behavior of LLMs well \cite{qin2024llmsmeetnlp,liu2024understandingllms,bowman2023eightthingsknow}.
    % 我们Single方法受限于分类性能（见表），因为预训练模型不能很好地预测大模型的行为。
    % \ourmethods is limited by classification performance (see Table~\ref{tab:classification}) because the Pre-trained Language Model (PLM) model cannot predict the behavior of LLMs well \cite{qin2024llmsmeetnlp,liu2024understandingllms,bowman2023eightthingsknow}.
    % 而受限于计算资源，我们没有尝试微调LLM进行分类
    However, we do not conduct experiments to fine-tune LLMs for classification limited by computing resources.
    % 我们将提升分类性能留作未来工作
    % We leave improving the classification performance for future work. 
    % 3. 我们的Vote方法也与oracle的性能有较大差距，因为我们使用的投票机制没有充分运用输出solutions中的丰富的语意信息
    (\emph{iii})~The performance of \ourmethodv also has a gap with the oracle performance, because the voting mechanism we adopt does not make full use of the rich semantic information in the LLM solutions, causing the limited improvement \cite{ni2023lever}.

\section{Cases Study}
    \label{app:case study}
\begin{figure*}
    \centering
    \includegraphics[width=1\linewidth]{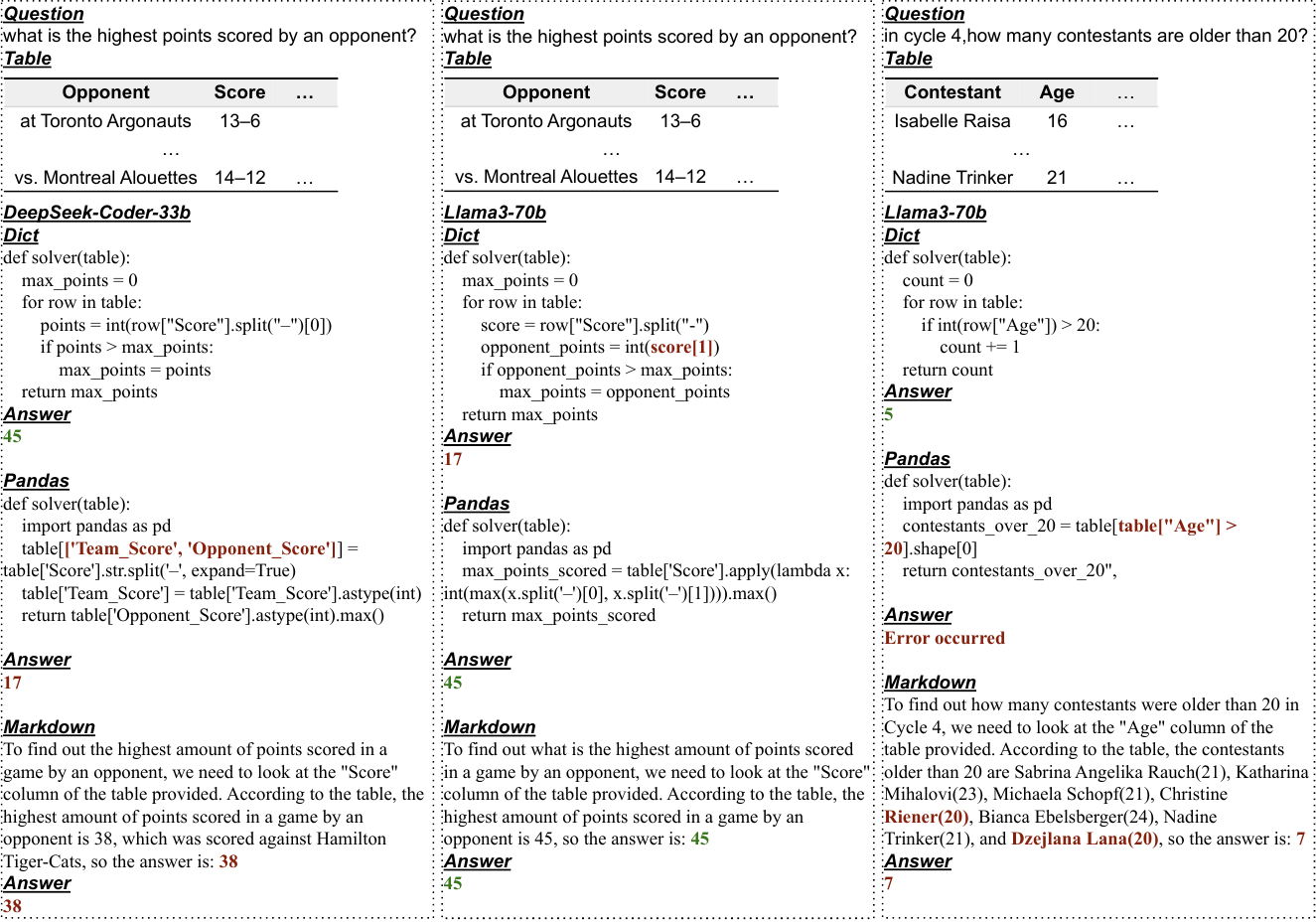}
    \caption{
    Two instances of WikiTQ test set using Llama3-70B and DeepSeek-Coder-33B with Markdown and Dict tabular format.
    }
    \label{fig:case_appendix}
\end{figure*}
    % 为了清楚展示表格表示的影响，我们在本节展示了更多WikiTQ中的例子，如图所示
    To clearly show the impact of the table, we show more instances in WikiTQ in this section, as shown in Figure~\ref{fig:case_appendix}.
    % 我们发现，
    We observe that:
    % 对于“what is the highest points scored by an opponent?”这个instance，DeepSeek-Coder-33B用Dict表示表格会正确求解，而Llama3-70B确适合用Markdown
    (\emph{i})~For the first instance with the question "\textit{what is the highest points scored by an opponent?}", DeepSeek-Coder-33B correctly solves the instance with the Dict format, while Llama3-70B is suitable for Markdown and Pandas formats.
    % 当我们使用相同的模型Llama3-70B时，在“in cycle 4,how many contestants are older than 20?”这个instance被用Dict正确求解，和上一个instance不同
    (\emph{ii})~When we use the same model Llama3-70B, the last instance with the question "\textit{in cycle 4, how many contestants are older than 20?}" is solved correctly with Dict, which is different from the previous instance.
    % 这也印证了我们的观点，instance有各自适合用于求解的表格表示，并且合适的表格表示还与用的LLM有关
    The cases claim that different instances and LLMs have different most suitable tabular formats.

\end{document}